\documentclass[11pt]{article}
\usepackage[table]{xcolor}
\usepackage{acl}
\usepackage{graphicx}
\usepackage{supertabular}
\usepackage{pgf}
\usepackage{amsmath}
\usepackage{fixltx2e}
\usepackage{times}
\usepackage{authblk}
\usepackage{latexsym}
\usepackage{multirow}
\usepackage{graphicx}
\usepackage{lscape}
\usepackage{longtable}
\usepackage{colortbl}
\usepackage{booktabs}
 \usepackage{multirow}
\usepackage{graphicx}
\usepackage[normalem]{ulem}

\usepackage{etoolbox}
\usepackage{pgf}
\usepackage{makecell}
\usepackage{xspace}

\usepackage[T1]{fontenc}

\usepackage[utf8]{inputenc}
\usepackage{silence}

\usepackage{microtype}

\usepackage{inconsolata}
\usepackage{amssymb}
\usepackage{pifont}

\usepackage{scalerel,graphicx,xparse}

\NewDocumentCommand\emojipoutingface{}{%
	\makebox[1em][l]{\raisebox{-0.1\height}{\includegraphics[height=1em,width=1em]{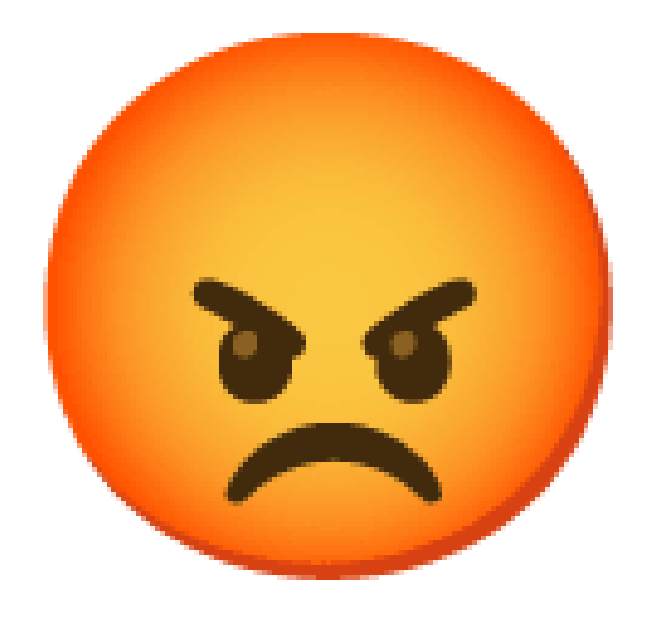}}}%
}

\NewDocumentCommand\emojigrinningface{}{%
	\makebox[1em][l]{\raisebox{-0.1\height}{\includegraphics[height=1em,width=1em]{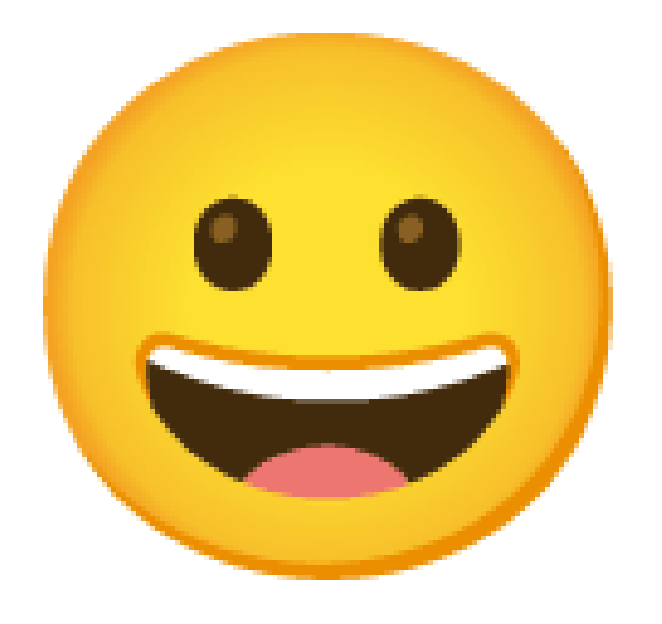}}}%
}

\NewDocumentCommand\emojibiohazard{}{%
	\makebox[1em][l]{\raisebox{-0.1\height}{\includegraphics[height=1em,width=1em]{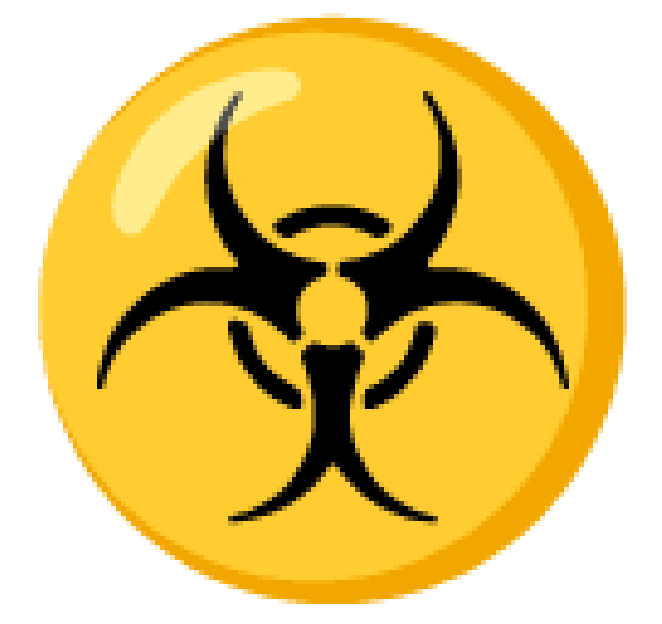}}}%
}

\NewDocumentCommand\emojipottedplant{}{%
	\makebox[1em][l]{\raisebox{-0.1\height}{\includegraphics[height=1em,width=1em]{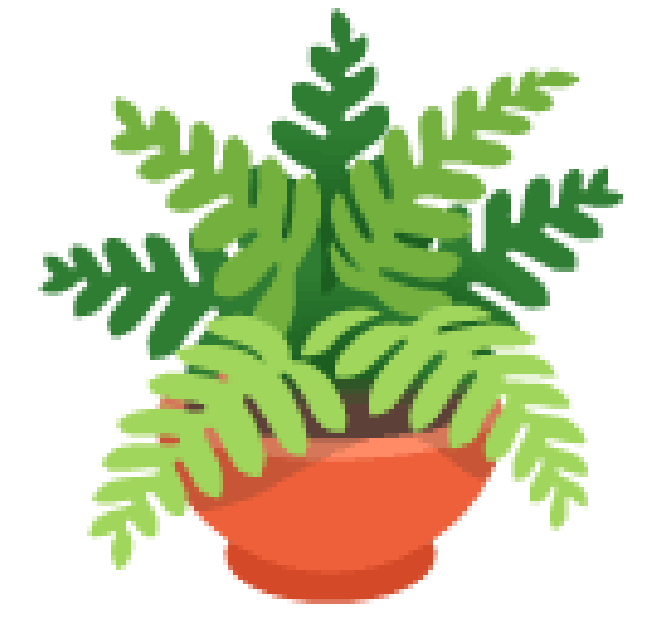}}}%
}

\renewcommand*{\Affilfont}{\normalsize\normalfont}

\definecolor{CarnegieRed}{HTML}{CC002B}
\definecolor{LeonieGreen}{HTML}{00883A}
\definecolor{LeonieBlue}{HTML}{00329D}
\definecolor{pastelblue}{rgb}{0.68, 0.85, 0.9}
\definecolor{pastelgreen}{rgb}{0.47, 0.87, 0.47}
\definecolor{pastelpink}{rgb}{1.0, 0.71, 0.76}
\definecolor{pastelpurple}{rgb}{0.7, 0.62, 0.71}
\definecolor{lightskyblue}{rgb}{0.81, 0.92, 0.97}
\definecolor{lightgreen}{rgb}{0.68, 0.93, 0.68}
\definecolor{lightpink}{rgb}{1.0, 0.88, 0.88}
\definecolor{lightpurple}{rgb}{0.85, 0.78, 0.91}
\definecolor{lightcoral}{rgb}{0.94, 0.5, 0.5}
\definecolor{lightcyan}{rgb}{0.88, 1.0, 1.0}
\definecolor{lightgoldenrodyellow}{rgb}{0.98, 0.98, 0.82}
\definecolor{lightgray}{rgb}{0.83, 0.83, 0.83}
\definecolor{lightseagreen}{rgb}{0.13, 0.7, 0.67}
\definecolor{lightsalmon}{rgb}{1.0, 0.63, 0.48}
\definecolor{lightsteelblue}{rgb}{0.69, 0.77, 0.87}
\definecolor{lightthistle}{rgb}{0.85, 0.75, 0.85}
\definecolor{lightpeach}{rgb}{1.0, 0.93, 0.87}
\definecolor{lightmint}{rgb}{0.96, 1.0, 0.98}
\definecolor{softbeige}{rgb}{0.96, 0.96, 0.86}
\definecolor{cream}{rgb}{1.0, 0.99, 0.82}
\definecolor{palelavender}{rgb}{0.9, 0.82, 0.95}
\definecolor{blushpink}{rgb}{1.0, 0.87, 0.9}
\definecolor{babyblue}{rgb}{0.84, 0.92, 0.98}
\definecolor{mintcream}{rgb}{0.96, 1.0, 0.98}
\definecolor{softlemon}{rgb}{0.98, 0.98, 0.88}
\definecolor{lightperiwinkle}{rgb}{0.8, 0.8, 0.96}
\definecolor{peachpuff}{rgb}{1.0, 0.85, 0.73}
\definecolor{lightsage}{rgb}{0.88, 0.93, 0.85}
\definecolor{lightteal}{rgb}{0.7, 0.9, 0.9}
\definecolor{lightolive}{rgb}{0.8, 0.8, 0.6}
\definecolor{lightmauve}{rgb}{0.86, 0.82, 0.91}
\definecolor{lightkhaki}{rgb}{0.94, 0.9, 0.55}
\definecolor{lightgold}{rgb}{0.98, 0.98, 0.82}
\definecolor{lightindigo}{rgb}{0.77, 0.7, 0.88}
\definecolor{lightmaroon}{rgb}{0.87, 0.72, 0.72}
\definecolor{lightsilver}{rgb}{0.85, 0.85, 0.85}
\definecolor{lightmoss}{rgb}{0.68, 0.87, 0.68}
\definecolor{lightazure}{rgb}{0.94, 1.0, 1.0}
\definecolor{apricot}{rgb}{0.98, 0.81, 0.69}
\definecolor{peach}{rgb}{1.0, 0.85, 0.72}
\definecolor{melon}{rgb}{0.99, 0.74, 0.71}
\definecolor{coral}{rgb}{1.0, 0.75, 0.6}
\definecolor{champagne}{rgb}{0.97, 0.91, 0.81}
\definecolor{lightsalmon}{rgb}{1.0, 0.63, 0.48}
\definecolor{cantaloupe}{rgb}{1.0, 0.8, 0.6}
\definecolor{atomictangerine}{rgb}{1.0, 0.6, 0.4}
\definecolor{pink}{rgb}{1.0, 0.75, 0.8}
\definecolor{rosepink}{rgb}{1.0, 0.63, 0.63}
\definecolor{carnationpink}{rgb}{1.0, 0.65, 0.79}
\definecolor{watermelon}{rgb}{0.99, 0.58, 0.59}
\definecolor{peachpink}{rgb}{1.0, 0.8, 0.7}
\definecolor{lightraspberry}{rgb}{0.98, 0.62, 0.69}
\definecolor{orchidpink}{rgb}{1.0, 0.74, 0.85}
\definecolor{petalpink}{rgb}{1.0, 0.91, 0.91}
\definecolor{gainsboro}{rgb}{0.86, 0.86, 0.86}
\definecolor{lightgray}{rgb}{0.83, 0.83, 0.83}
\definecolor{silver}{rgb}{0.75, 0.75, 0.75}
\definecolor{platinum}{rgb}{0.9, 0.89, 0.89}
\definecolor{ashgray}{rgb}{0.7, 0.75, 0.71}
\definecolor{timberwolf}{rgb}{0.86, 0.84, 0.82}
\definecolor{smokegray}{rgb}{0.71, 0.74, 0.76}
\definecolor{coolgray}{rgb}{0.55, 0.57, 0.67}
\definecolor{bluegray}{rgb}{0.67, 0.75, 0.77}
\definecolor{palesilver}{rgb}{0.79, 0.75, 0.73}

\makeatletter
\def\gcmidrule{\arrayrulecolor{lightgray}%
    \noalign{\ifnum0=`}\fi
    \@ifnextchar[{\@gcmidrule}{\@gcmidrule[\cmidrulewidth]}}
\def\@gcmidrule[#1]{\@ifnextchar({\@@gcmidrule[#1]}{\@@gcmidrule[#1]()}}
\def\@@gcmidrule[#1](#2)#3{\@@@gcmidrule[#3]{#1}{#2}}
\def\@@@gcmidrule[#1-#2]#3#4{\global\@cmidla#1\relax
    \global\advance\@cmidla\m@ne
    \ifnum\@cmidla>0\global\let\@gtempa\@cmidrulea\else
    \global\let\@gtempa\@cmidruleb\fi
    \global\@cmidlb#2\relax
    \global\advance\@cmidlb-\@cmidla
    \global\@thisrulewidth=#3
    \@setrulekerning{#4}
    \ifnum\@lastruleclass=\z@\vskip \aboverulesep\fi
    \ifnum0=`{\fi}\@gtempa
    \noalign{\ifnum0=`}\fi\futurenonspacelet\@tempa\@xgcmidrule}
\def\@xgcmidrule{%
   \ifx\@tempa\gcmidrule
       \vskip-\@thisrulewidth
       \global\@lastruleclass=\@ne
   \else \ifx\@tempa\morecmidrules
       \vskip \cmidrulesep
       \global\@lastruleclass=\@ne\else
       \vskip \belowrulesep
       \global\@lastruleclass=\z@
   \fi\fi
   \ifnum0=`{\fi}
  \arrayrulecolor{black}}%
\makeatother

\definecolor{high}{HTML}{689F38}  %
\definecolor{low}{HTML}{D32F2F}  %
\newcommand*{\opacity}{40}%
\newcommand*{\minval}{0.0}%
\newcommand*{\maxval}{100.0}%
\newcommand{\fixedwidthbox}[1]{%
  \makebox[0.7cm][c]{#1}  %
}

\newcommand{\gradient}[1]{
    \ifdimcomp{#1pt}{>}{\maxval pt}{\fixedwidthbox{#1}}{
        \ifdimcomp{#1pt}{<}{\minval pt}{\fixedwidthbox{#1}}{
            \pgfmathparse{int(round(100*(#1/(\maxval-\minval))-(\minval*(100/(\maxval-\minval)))))}
            \xdef\tempa{\pgfmathresult}
            \cellcolor{high!\tempa!low!\opacity}\fixedwidthbox{#1}
        }
    }
}

\title{\framework: Hybrid Behavioral Testing of NLP Models \\with Synthetic CheckLists\\
\textit{\textcolor{CarnegieRed}{\scalebox{0.6}{Warning: This
      paper contains language that readers may find offensive or disturbing.}}}}

\makeatletter
\renewcommand\AB@affilsepx{\hspace{1em} \protect\Affilfont}
\makeatother

\author[1,2,3]{Raoyuan Zhao}
\author[2,3,4]{Abdullatif Köksal}
\author[2,3]{Yihong Liu}
\author[2,3]{Leonie Weissweiler}
\author[4]{\authorcr Anna Korhonen}
\author[2,3]{Hinrich Schütze}

\affil[1]{Technical University of Munich}
\affil[2]{LMU Munich\protect\\\hspace{-0.5cm}}
\affil[3]{Munich Center for Machine Learning}
\affil[4]{University of Cambridge \protect\\
	\texttt{raoyuan.zhao@tum.de, akoksal@cis.lmu.de}}

\newcommand{\framework}{\textsc{SynthEval}\xspace}
\newcommand{\taskmodel}{TaskModel\xspace}
\newcommand{\taskmodels}{TaskModels\xspace}

\begin{document}
\maketitle
\begin{abstract}

Traditional benchmarking in NLP typically involves using
static held-out test sets.
However, this approach
often results in an overestimation of performance and lacks
the ability to offer comprehensive, interpretable, and
dynamic assessments of NLP models.  Recently, works like
DynaBench \citep{kiela-etal-2021-dynabench} and CheckList
\citep{ribeiro-etal-2020-beyond} have addressed these
limitations through \emph{behavioral testing} of NLP models
with \emph{test types} generated by a multi-step
human-annotated pipeline. Unfortunately, manually creating a
variety of test types requires much human labor,
often at prohibitive cost.
In this work, we propose
\textbf{\framework}, a hybrid behavioral testing framework
that leverages large language models (LLMs) to generate a
wide range of test types for a comprehensive evaluation of
NLP models.  \framework first generates
sentences via LLMs using controlled generation, and then
identifies challenging examples by comparing the predictions
made by LLMs with task-specific NLP models. In the last
stage, human experts investigate the challenging examples,
manually design templates, and identify the types of
failures the task-specific models consistently exhibit. We
apply \framework to two classification tasks,
sentiment analysis and toxic language detection,
and show that
our framework is effective in identifying weaknesses of
strong models on these tasks. We share our code in \url{https://github.com/Loreley99/SynthEval_CheckList}.

\end{abstract}

\section{Introduction}

\begin{figure}[t]
	\centering
	\includegraphics[width=0.44\textwidth]{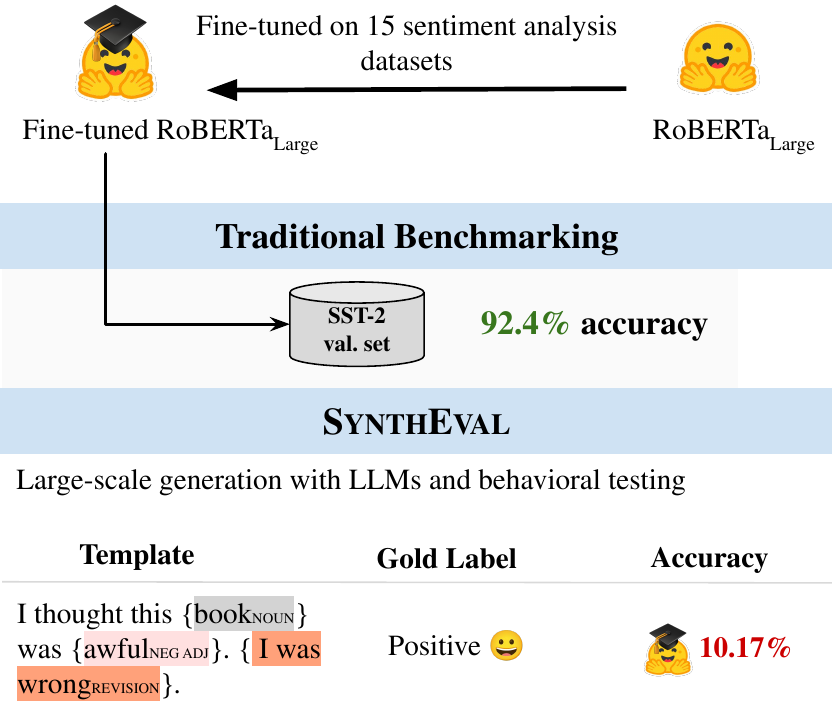}
	\caption{Using a held-out val. set for evaluation
		overestimates the
		performance. RoBERTa\textsubscript{Large}, fine-tuned on
		15 diverse sentiment analysis datasets
		\cite{hartmann2023},
		performs strongly on traditional benchmarks (92.4\%). However, \framework, which generates behavioral
		tests with the help of LLMs, demonstrates
		RoBERTa\textsubscript{Large}'s bad performance (10.17\%)
		when tested on a sentence containing a simple revision: ``I was wrong''.}
	\label{fig:first_page_figure_emnlp}
\end{figure}

The typical pipeline before deploying NLP models for
practical use involves training, validating, and testing
phases. A model that performs well on a held-out test set,
as measured by a single aggregate statistic, e.g., accuracy,
is expected to be capable of generalization
\citep{roelofs2019measuring}. Unfortunately, such a measure
often leads to an overestimation of real-world performance,
as validation and test sets are likely to contain similar
biases as the train set
\citep{Torralba2011dataset,rudinger-etal-2017-social,
  rajpurkar-etal-2018-know}. In Figure
\ref{fig:first_page_figure_emnlp}, we see a fine-tuned RoBERTa
model evaluated on the test set of SST-2
\citep{socher-etal-2013-recursive}.
It achieves high accuracy (92.4\%), but
fails (only 10.17\% accuracy) when
a revision like \texttt{I was wrong.} is appended to
a simple sentence on which the model can correctly
identify the sentiment.

Despite the broad capabilities of large language models (LLMs), their 
immense computational and resource requirements often render them 
impractical for deployment in scenarios with limited infrastructure or 
for low-resource languages \cite{naveed2023comprehensive}. 
Also, training and deploying LLMs demands considerable computational power
and data, raising significant economic and environmental concerns \cite{strubell2020energy,patterson2021carbon}. 
In contrast,
we are interested in
task-specific NLP models, which we refer to as 
\textbf{\taskmodels} in this paper -- 
task-specific pre-trained language models (PLMs) such as BERT, DistilBERT and RoBERTa that are
fine-tuned on labeled data for specific NLP tasks such as text
classification, named entity recognition and part-of-speech
tagging.
\taskmodels have the advantage of being compact, efficient
and effective. These qualities make \taskmodels highly
relevant -- even in the era of LLMs --
and they are widely used in resource-constrained environments.

To assess the true capabilities of a \taskmodel, especially
uncovering its \textit{vulnerabilities}, many works go beyond
simply evaluating against a single aggregate
statistic. These approaches evaluate multiple aspects of a
model such as robustness, consistency and error
types 
 \citep{Belinkov2018noise,ribeiro-etal-2019-red,wu-etal-2019-errudite,gardner-etal-2020-evaluating}. Though providing ways to evaluate a model's competence in different facets, these methods fail to provide comprehensive guidance on how to evaluate the model. CheckList, a method that breaks down capability failures into specific failures, is proposed to fill the gap \citep{ribeiro-etal-2020-beyond}. CheckList leverages a strategy called \emph{behavioral testing} or \emph{black-box testing},  originating from software engineering, which tests the application by providing inputs and then examining the outputs without knowing what the software does internally to arrive at those outputs \citep{Beizer1995blackbox}.

Although recent works like CheckList
\citep{ribeiro-etal-2020-beyond} and DynaBench
\cite{kiela-etal-2021-dynabench} introduce a wide range of
\emph{test types} by proposing multi-step human-annotated
evaluation and template-based analysis of NLP models, these
types are manually extracted and summarized. This involves
substantial human labor, which is not only tedious but can
also miss model vulnerabilities.
Due to recent advancement of
LLMs, they now have a human-comparable ability to
generate many types of high-quality data \citep{koksal2023hallucination,
  ye-etal-2022-zerogen,whitehouse-etal-2023-llm,yu2023llmgen,heng2024proggen}. Because
of this capability, LLMs have also been used to help
identify and even fix the possible weaknesses of smaller
models, either through LLMs suggesting the test types
\citep{ribeiro-lundberg-2022-adaptive} or through LLMs
generating instances for test types
\citep{he-etal-2023-targeted}. These methods
either heavily require human effort in the pipeline, or ask
for the test types to be extracted before LLMs can
generate instances for them. Inspired by this line of work, we pose two
research questions: (1) How can we directly generate a large
number of test types using LLMs? and (2) How can we reduce
the burden of annotators when identifying challenging test
types?

To this end, we propose \framework, a novel hybrid
behavioral testing methodology based on LLMs. \framework
leverages an LLM to generate diverse examples for a given
task (in our case, classification). The generated examples
are fed to both a \taskmodel (the model that we
perform behavioral testing on) and the reference model (the
same LLM used to generate the examples) for
predictions. Then human experts need only extract and
summarize the examples on which the prediction of \taskmodel and reference model diverge. Finally, we can generate a lot of test types automatically by applying the templates created by human experts. The procedure greatly reduces human labor and enables us to generate diverse examples. 

The contributions of this work are as follows: (i) We
propose \framework, a framework that
partially automates the process of generating diverse and
challenging test types for evaluating NLP models. (ii) We
validate the reliability of \framework on two classification
tasks: sentiment analysis and toxic language
detection. (iii) We conduct a comprehensive, linguistically informed analysis to identify patterns in sentences where classification models face challenges.

\section{Related Work}

\paragraph{Synthetic Datasets} Traditional classification datasets often originate from texts written and labeled by humans \cite{founta2018large,farber2020multidimensional,kennedy2018gab}. However, recent studies have highlighted the feasibility of machine-generated synthetic datasets \cite{Trinh2024}, showing LLMs' ability to produce texts comparable to human writing \cite{jawahar-etal-2020-automatic,clark-etal-2021-thats}. Additionally, LLMs potentially generate data that is more diverse and comprehensive than what is typically possible through human efforts \cite{munoz2023contrasting, hartvigsen-etal-2022-toxigen}. Recent works explore hybrid synthetic datasets via LLMs and corpora for instruction tuning \cite{koksal2024longform} and evaluating rare linguistic phenomena \cite{weissweiler2024hybrid}.

\paragraph{Limitations in Handling Linguistic Complexity}
While classification models, including those used for tasks
like sentiment analysis, have demonstrated the ability to
outperform humans on traditional datasets
\cite{dang2020sentiment}, they face limitations with more
complex linguistic structures. Recent studies indicate that
language models, despite their advancements, often struggle
with complex syntax and nuanced
expressions. \citet{rogers-etal-2020-primer} reveal that
neural language models like BERT still struggle with complex
syntactic sentences easily handled by humans. Experiments
conducted by \citet{kassner-schutze-2020-negated}
demonstrate that pretrained language models often fail to
correctly process  negation in sentences. \citet{hall-maudslay-cotterell-2021-syntactic} also indicate that language models like BERT, GPT-2, and RoBERTa rely on semantic cues for syntactic predictions, highlighting a limitation in their syntactic understanding.

\paragraph{Interpretable Behavioral Testing}
\citet{ribeiro-etal-2020-beyond} propose CheckList, a suite
of tests for evaluating model robustness across various
linguistic phenomena. This framework helps pinpoint
fundamental linguistic shortcomings in models that perform
well on standard benchmarks. However, CheckList is
challenging in
practice: it is costly and requires experienced annotators who are competent to conduct
qualitative assessments of templates
\cite{lee2024checkeval,k-etal-2022-multilingual}. \citet{yang-etal-2022-testaug} propose TestAug, which utilizes GPT-3 to generate more test cases based on CheckList's existed templates, but no new patterns are detected. \citet{ferrando-etal-2023-automating} also attempt to use LLM to try to reduce the human burden of testing the performance of machine translation systems. 
Other methodologies like HATECHECK \cite{rottger-etal-2021-hatecheck}, Red Teaming \cite{perez-etal-2022-red}, and Targeted Data Generation \cite{he-etal-2023-targeted} complement CheckList by offering dynamic ways to test and improve model evaluation. These approaches underscore the need for ongoing, refined assessments to build more robust NLP systems.
Recent developments in these areas focus on scalability, enhancing template quality, and ensuring relevance across various languages and cultures. Adaptive Testing \cite{ribeiro-lundberg-2022-adaptive} highlights the benefits of integrating human insights with LLMs in testing frameworks, pointing to a more collaborative approach in advancing model reliability and fairness.

Prior methods require substantial human intervention for creating templates and identifying confusing sentences without any references, which can be labor-intensive and may overlook issues.

\section{Methodology: \framework}
We propose \framework, a hybrid and dynamic evaluation
framework to reveal behavioral failures of task-specific NLP
models
with the help of large language models and human
annotation.
We refer to task-specific NLP
models as \taskmodels;
\taskmodels are pretrained language
models finetuned to perform a specific task such as sentiment analysis classification. These models are widely used in industry, even in the era of large language models, since they achieve good performance on test sets and are cheaper to deploy.

\begin{figure*}[htb]
    \setlength{\belowcaptionskip}{-0.5cm}
\centering
\includegraphics[width=0.92\textwidth]{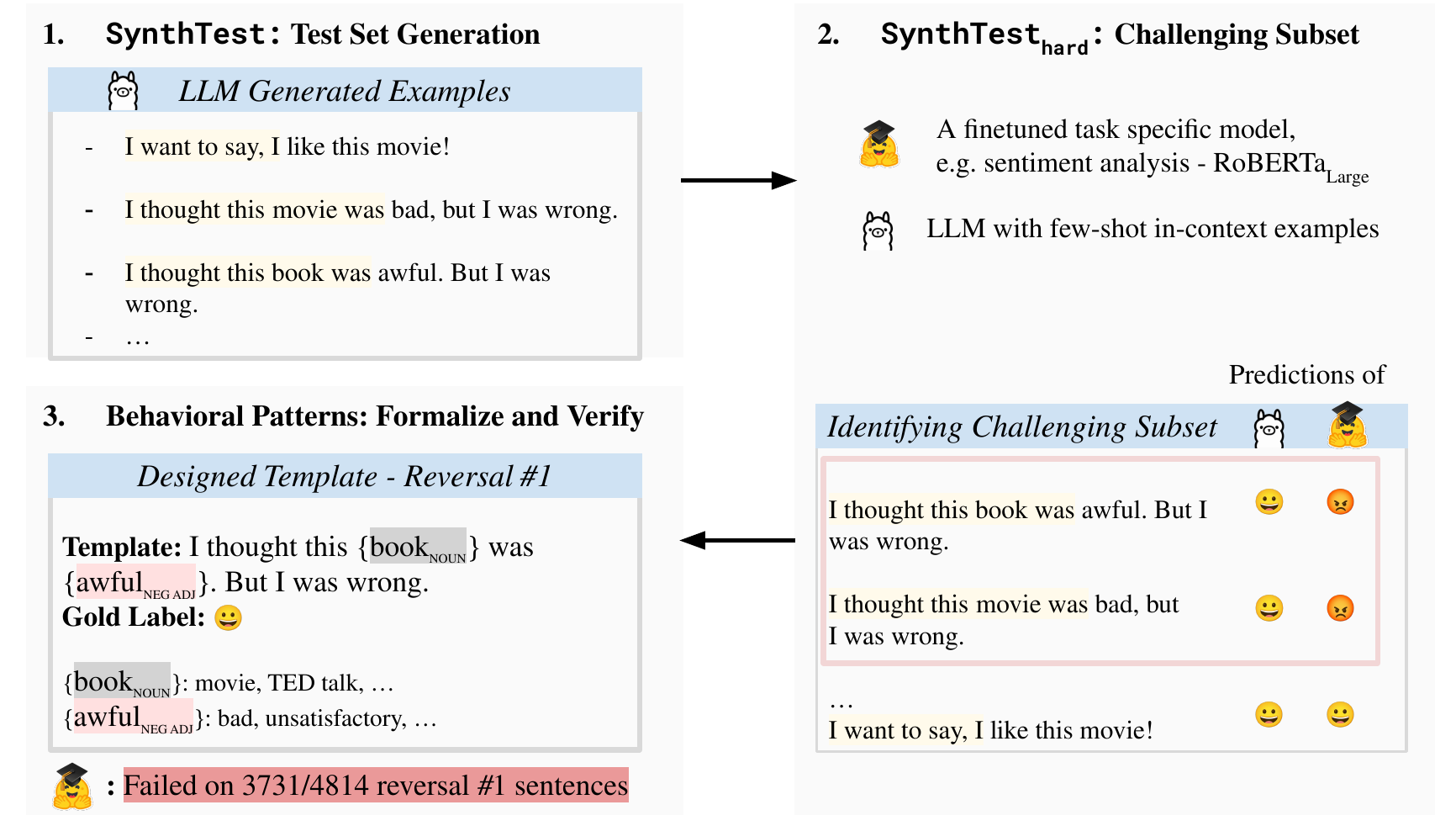}
\caption{A summary of \framework with the sentiment analysis task as an example. It consists of three steps: 1. Generating a diverse and large-scale test set with LLMs. 2. Identifying a challenging subset by comparing predictions between a \taskmodel (i.e., RoBERTa) and a reference model (i.e., few-shot LLM), and sorting based on differences. 3. Manually designing behavioral patterns and evaluating \taskmodels accordingly.}

\label{fig:workflow}
\end{figure*}

In Figure \ref{fig:workflow}, we illustrate \framework, which consists of three steps: (1) Diverse synthetic test set generation (\texttt{SynthTest}), (2) Identification of a challenging subset of the test set (\texttt{SynthTest\textsubscript{hard}}), and (3) Manual formalization and verification of behavioral patterns from \texttt{SynthTest\textsubscript{hard}}.

\subsection{\texttt{SynthTest}: Test Set Generation}

In Figure \ref{fig:workflow}, we can see that the first stage of \framework involves generating a diverse test set without gold labels. By gathering a large-scale test set, we aim to cover a wide range of test cases, thereby revealing potential limitations of \taskmodels.

We leverage LLMs to generate sentences under certain constraints as they can generate a diverse range of sentences at a lower cost \cite{hartvigsen-etal-2022-toxigen}. First, we randomly sample words (queries) from existing datasets to guide generation via LLMs, as zero-shot synthetic data generation from LLMs tends to be less diverse \cite{li-etal-2023-synthetic}. As illustrated in Figure \ref{fig:workflow}, we randomly sample 5 words from existing datasets and prompt the LLM to continue the text.

We employ nucleus sampling \cite{holtzman2019curious} with $p=1.0$ to increase diversity during this step and generate 100,000 test set examples for each task. Since the LLM does not indicate the end of the sentence, we use an additional sentence segmentation model to extract only the first sentence.

\subsection{\texttt{SynthTest\textsubscript{hard}}: Challenging Subset}
In the second stage of \framework, we aim to identify challenging examples for the \taskmodel. By identifying these examples, we aim to increase \framework's automation and reduce the workload compared to \citet{ribeiro-lundberg-2022-adaptive}.

As we do not have gold labels for \texttt{SynthTest}, we
cannot find challenging examples by simply comparing the
predictions of the \taskmodel with gold labels. Therefore,
we adopt an approach similar to ensembling to
identify challenging examples. We also make predictions
using
LLMs with few-shot in-context learning, which typically achieves
comparable performance to \taskmodels but better
generalization capabilities \cite{NEURIPS2020_1457c0d6}. We
use these predictions as an additional signal to find
the most challenging examples. Specifically, we calculate
the absolute difference between the probability of the most
likely label from the \taskmodels and the
probability of the same label from the LLM. Then, we sort
the examples in \texttt{SynthTest} by the absolute
difference of prediction probabilities and focus on the
first 10,000 examples, which we refer to as \texttt{SynthTest\textsubscript{hard}}. As illustrated in Figure \ref{fig:workflow}, sorting examples by the absolute difference reveals more challenging examples for the \taskmodel.

\subsection{Behavioral Patterns: Formalize and Verify}
The last stage of \framework involves finding consistent behavioral patterns that cause failures in the \taskmodel. For this purpose, we employ both automated analysis and manual analysis via human annotators.

The first step is a manual investigation to identify
examples in \texttt{SynthTest\textsubscript{hard}} for which
\taskmodel struggles to make accurate
predictions. We first extract the most frequent n-grams,
aiming to identify specific words or phrases that the
\taskmodel appears to not fully comprehend. We
manually investigate the specific examples within the same
n-gram groups and examine if there are any systematic
errors. For example, this analysis reveals that sentences
including the phrase ``was blown away'' are consistently
interpreted incorrectly by the \taskmodel\ -- even
though it achieves over 92\% accuracy on standard benchmarks. We also manually investigate sentences separately, especially for the examples with the largest absolute difference.

The second step is formalizing behavioral patterns from
these manually filtered examples and creating simpler test
sentences with the same label set. For this purpose, we
hypothesize why each of these failures occurs, with the help
of n-gram frequencies and manual analysis, and develop
simple sentences. We create placeholders for lexical groups
in those sentences, such as nouns, positive/negative
adjectives, nationality groups, and typos. Thus, we end up
with behavioral template patterns that can be populated to
hundreds of sentences with the same label.
Figure~\ref{fig:workflow} illustrates this
with the template ``Designed Template -- Reversal \#1'', which
consists of 
the behavioral pattern
\texttt{I thought this \{NOUN\} was \{NEG ADJ\}. But
  I was wrong.}; it has the ground truth label (positive). A
list of nouns and positive adjectives that can be
substituted is shown.

The final step is the verification of the behavioral
patterns. Since we now know the ground truth of the
generated sentences, the \taskmodel's accuracy can
be evaluated on them. For example, in Figure
\ref{fig:workflow}, the \taskmodel fails on 3731 out
of 4814 generated sentences from the Reversal \#1 template. This
verifies that the \taskmodel fails to understand
this specific phrasing.
Thus our \framework methodology
enables a more interpretable evaluation of the
\taskmodel\
-- and an evaluation that reveals critical failures
that are concealed by good performance on
traditional benchmarks.

\section{Analysis}

We apply \framework on two diverse tasks: sentiment
analysis and text toxicity detection. We selected them 
since they are
widely studied both in academia and industry. Furthermore, current \taskmodels for
sentiment analysis can achieve very high performance on
traditional benchmarks which presents an interesting challenge for \framework to uncover potential weaknesses.
 Of the two, toxicity detection is the more
challenging task since it requires recognizing
harmful content that is often implied and
context-sensitive. Toxicity detection failure
of current models
can 
marginalize  minority groups
\cite{sap-etal-2019-risk, diaz2021double}.
Thus, more effective
behavioral
testing has great potential benefits for this task.

In our experiments, we use LLaMA2 7B \cite{touvron2023llama}
as the main LLM both for test set generation and for creating a
challenging subset with few-shot in-context learning.

\subsection{Sentiment Analysis}
We choose two \taskmodels for sentiment analysis
with two classes: positive and negative. The first one is
SiEBERT \cite{hartmann2023}, a strong RoBERTa large model
\cite{liu2019roberta} that is fine-tuned on 15 diverse
sentiment analysis tasks \cite{liu2019roberta}. The second
model is DistilBERT \cite{sanh2019distilbert} fine-tuned on
SST-2 \cite{socher-etal-2013-recursive}. When we test these
models on a traditional benchmark, the validation set of
SST-2, both models achieve strong performance: 92.4\%
accuracy for RoBERTa large  and 91.0\% for
DistilBERT. As a reference model in the second step of
\framework, we use 4-shot LLaMA2 7B; its accuracy is
93.0\%.

During the application of \framework, we randomly sample 100,000 sentences from the IMDb training dataset and select the top five words of each sentence as queries for generating new sentences. We then use 4-shot LLaMA2 7B and two \taskmodels, SiEBERT, and finetuned DistilBERT to classify these generated sentences separately. Next, we compute the probability difference of the positive label between the reference model and each \taskmodel separately, and annotators analyze the top 10,000 sentences with the highest probability difference. This process allows us to identify and summarize a total of 12 challenging patterns, which are detailed in Table \ref{result_sentiment}. We now highlight several linguistic factors that cause consistent failures in our \taskmodels for sentiment analysis.

\begin{table*}[h!]
\centering
\resizebox{\linewidth}{!}{%
\begin{tabular}{llrrc}

\toprule
\textbf{Test Type} & \textbf{Template} & \multicolumn{2}{c}{\textbf{Accuracy(\%)}} & \textbf{Gold} \\ 
&&\textbf{DistilBERT} & \textbf{RoBERTa} & \textbf{Label} \\ \midrule
\multirow{15}{*}{\textbf{Negation}} & \scalebox{1}{This \{\colorbox{lightgray}{book\textsubscript{\scalebox{0.50}{NOUN}}}\}}  \scalebox{1}{is not} $\left\{\begin{array}{l} \colorbox{lightpink}{awful\textsubscript{\scalebox{0.5}{NEG ADJ}}} \\ \colorbox{lightgreen}{nice\textsubscript{\scalebox{0.5}{POS ADJ}}}\end{array} \right\}$. &  $\begin{array}{r} \gradient{99.79} \\ \gradient{100.0}\end{array}$ & $\begin{array}{r} \gradient{100.0} \\ \gradient{99.43}\end{array}$ & $\begin{array}{r} \text{\emojigrinningface} \\
\text{\emojipoutingface}\end{array}$\\   
\gcmidrule(lr){2-2}
& I don't think this \{\colorbox{lightgray}{book\textsubscript{\scalebox{0.50}{NOUN}}}\} is $\left\{\begin{array}{l} \colorbox{lightpink}{awful\textsubscript{\scalebox{0.5}{NEG ADJ}}} \\ \colorbox{lightgreen}{nice\textsubscript{\scalebox{0.5}{POS ADJ}}}\end{array}\right\}$. & $\begin{array}{r} \gradient{46.92} \\ \gradient{99.40}\end{array}$ & $\begin{array}{r} \gradient{70.52} \\ \gradient{100.0}\end{array}$ & $\begin{array}{r} \text{\emojigrinningface} \\ \text{\emojipoutingface}\end{array}$\\
\gcmidrule(lr){2-2}
& It isn't true that this \{\colorbox{lightgray}{book\textsubscript{\scalebox{0.50}{NOUN}}}\} isn't $\left\{\begin{array}{l} \colorbox{lightgreen}{nice\textsubscript{\scalebox{0.5}{POS ADJ}}} \\ \colorbox{lightpink}{awful\textsubscript{\scalebox{0.5}{POS ADJ}}}\end{array}\right\}$. & $\begin{array}{r} \gradient{0.00} \\ \gradient{35.08}\end{array}$ & $\begin{array}{r} \gradient{76.87} \\ \gradient{97.52}\end{array}$ & $\begin{array}{r} \text{\emojigrinningface} \\ \text{\emojipoutingface}\end{array}$ \\
\gcmidrule(lr){2-2}
& I can't find anything$\left\{\begin{array}{l} \colorbox{lightpink}{awful\textsubscript{\scalebox{0.5}{NEG ADJ}}} \\ \colorbox{lightgreen}{nice\textsubscript{\scalebox{0.5}{POS ADJ}}}\end{array}\right\}$ to say about this \{\colorbox{lightgray}{book\textsubscript{\scalebox{0.50}{NOUN}}}\}. & $\begin{array}{r} \gradient{5.95} \\ \gradient{88.65}\end{array}$ & $\begin{array}{r} \gradient{91.50} \\ \gradient{99.20}\end{array}$ & $\begin{array}{r} \text{\emojigrinningface} \\ \text{\emojipoutingface}\end{array}$\\
\gcmidrule(lr){2-2}
& I am unable to find anything$\left\{\begin{array}{l} \colorbox{lightpink}{awful\textsubscript{\scalebox{0.5}{NEG ADJ}}} \\ \colorbox{lightgreen}{nice\textsubscript{\scalebox{0.5}{POS ADJ}}}\end{array}\right\}$to say about this \{\colorbox{lightgray}{book\textsubscript{\scalebox{0.50}{NOUN}}}\}. & $\begin{array}{r} \gradient{5.88} \\ \gradient{99.93}\end{array}$ & $\begin{array}{r} \gradient{86.61} \\ \gradient{98.90}\end{array}$ & $\begin{array}{r} \text{\emojigrinningface} \\ \text{\emojipoutingface}\end{array}$\\
\gcmidrule(lr){2-2}
& I don't find anything$\left\{\begin{array}{l} \colorbox{lightpink}{awful\textsubscript{\scalebox{0.5}{NEG ADJ}}} \\ \colorbox{lightgreen}{nice\textsubscript{\scalebox{0.5}{POS ADJ}}}\end{array}\right\}$to say about this \{\colorbox{lightgray}{book\textsubscript{\scalebox{0.50}{NOUN}}}\}. & $\begin{array}{r} \gradient{38.27} \\ \gradient{95.04}\end{array}$ & $\begin{array}{r} \gradient{100.00} \\ \gradient{99.77}\end{array}$ & $\begin{array}{r} \text{\emojigrinningface} \\ \text{\emojipoutingface}\end{array}$\\  
\midrule
\multirow{4}{*}{\textbf{Past Tense}}  
& \makecell{\{\colorbox{lightsalmon}{\text{I was wrong.}\textsubscript{\scalebox{0.5}{REVISION}}}\} I thought this \{\colorbox{lightgray}{book\textsubscript{\scalebox{0.50}{NOUN}}}\} was $\left\{\begin{array}{l} \colorbox{lightpink}{awful\textsubscript{\scalebox{0.5}{NEG ADJ}}} \\ \colorbox{lightgreen}{nice\textsubscript{\scalebox{0.5}{POS ADJ}}}\end{array}\right\}$.} & $\begin{array}{r} \gradient{0.00} \\ \gradient{6.03}\end{array}$ & $\begin{array}{r} \gradient{0.56} \\ \gradient{10.86}\end{array}$& $\begin{array}{r} \text{\emojigrinningface} \\ \text{\emojipoutingface}\end{array}$\\  
\gcmidrule(lr){2-2}
& \makecell{I thought this \{\colorbox{lightgray}{book\textsubscript{\scalebox{0.50}{NOUN}}}\} was $\left\{\begin{array}{l} \colorbox{lightpink}{awful\textsubscript{\scalebox{0.5}{NEG ADJ}}} \\ \colorbox{lightgreen}{nice\textsubscript{\scalebox{0.5}{POS ADJ}}}\end{array}\right\}$. \{\colorbox{lightsalmon}{\text{I was wrong.}\textsubscript{\scalebox{0.5}{REVISION}}}\}} & $\begin{array}{r} \gradient{2.78} \\ \gradient{44.84}\end{array}$ & $\begin{array}{r} \gradient{10.17} \\ \gradient{57.29}\end{array}$ & $\begin{array}{r} \text{\emojigrinningface} \\ \text{\emojipoutingface}\end{array}$ \\  
\midrule
\multirow{4}{*}{\textbf{Comparative}} & \makecell{I'm sure I'll see plenty in the future, but I'm sure none will be as $\left\{\begin{array}{l} \colorbox{lightgreen}{nice\textsubscript{\scalebox{0.5}{POS ADJ}}} \\ \colorbox{lightpink}{awful\textsubscript{\scalebox{0.5}{NEG ADJ}}}\end{array}\right\}$ \\as this \{\colorbox{lightgray}{book\textsubscript{\scalebox{0.50}{NOUN}}}\}.} & $\begin{array}{r} \gradient{0.67} \\ \gradient{98.65}\end{array}$ & $\begin{array}{r} \gradient{100.00} \\ \gradient{100.00}\end{array}$ & $\begin{array}{r} \text{\emojigrinningface} \\ \text{\emojipoutingface}\end{array}$\\  
\gcmidrule(lr){2-2}
& There can't be any $\left\{\begin{array}{l} \colorbox{lightcoral}{\text{worse}\textsubscript{\scalebox{0.5}{NEG COMPARATIVE ADJ}}} \\ \colorbox{lightsage}{\text{better}\textsubscript{\scalebox{0.5}{POS COMPARATIVE ADJ}}}\end{array}\right\}\textsubscript{\scalebox{0.75}{}}$ \{\colorbox{lightgray}{book\textsubscript{\scalebox{0.50}{NOUN}}}\} than this one. & $\begin{array}{r} \gradient{78.31} \\ \gradient{0.00}\end{array}$ & $\begin{array}{r} \gradient{99.20} \\ \gradient{100.00}\end{array}$& $\begin{array}{r}  \text{\emojipoutingface} \\ \text{\emojigrinningface}\end{array}$  \\  
\midrule
& I was blown away by this \{\colorbox{lightgray}{book\textsubscript{\scalebox{0.50}{NOUN}}}\}. & $\begin{array}{r}\gradient{0.00} \end{array}$& $\begin{array}{r} \gradient{98.80}  \end{array}$& $\begin{array}{r} \text{\emojigrinningface} \end{array}$\\  
\gcmidrule(lr){2-2}
\multirow{-2}{*}{\textbf{Specific Phrase}} & This \{\colorbox{lightgray}{book\textsubscript{\scalebox{0.50}{NOUN}}}\} is a perfect little atrocity... & $\begin{array}{r} \gradient{0.00} \end{array}$& $\begin{array}{r} \gradient{0.00}  \end{array}$& $\begin{array}{r} \text{\emojipoutingface} \end{array}$ \\ 
\midrule
\end{tabular}%
}
\caption{\framework templates for sentiment analysis and
  accuracy for two \taskmodels. Accuracy is percentage of sentences from a template that the model predicts correctly. \text{\emojipoutingface} represents the sentiment is negative, while \text{\emojigrinningface} means the sentiment is positive.}
\label{result_sentiment}
\end{table*}

\paragraph{Negation}
Table \ref{result_sentiment} confirms 
the well-attested finding
\cite{kassner-schutze-2020-negated} that negation is a
challenge for smaller language models. For example, 
for the sentence template \texttt{I \textbf{don't} think this
  \{NOUN\} is \{NEG ADJ\}. }, both models perform
poorly, with accuracies of only 46.92\% and 70.52\%.
The more complex
double negative template \texttt{It \textbf{isn't} true
  that this \{NOUN\} \textbf{isn't} \{POS ADJ\}. }
reduces DistilBERT's accuracy to 0\% while RoBERTa gets almost a quarter wrong (accuracy of 76.87\%).

\paragraph{Past Tense}
This test type 
covers
statements about the subject
in the past tense that are then revised. For instance, when
testing with
the ``negative-to-positive revision''
template \texttt{I \textbf{thought} this
  \{NOUN\} was \{NEG ADJ\}. \{REVISION\}},
the accuracy of both models is low. Models particularly
struggle
with this negative-to-positive revision.
In contrast, in
the ``positive-to-negative revision'' template
\texttt{I \textbf{thought} this
  \{NOUN\} is \{POS ADJ\}. \{REVISION\}},
both models are highly accurate.
The reason could be that movie critics more often use
positive-to-negative revision and therefore it is better
represented in training data.

\paragraph{Order}
The order of words and lingustic constituents also has a large impact on model
comprehension. For instance, in the Past Tense test type,
moving
\texttt{\{REVISION\}} in the template \texttt{\texttt{I
    thought this \{NOUN\} was \{POS ADJ\}.\{REVISION\}}} from the end to the beginning
drastically reduces DistilBERT's accuracy
from 44.84\% to 6.03\%. This could be due to common language patterns where revisions or contrasting statements typically appear at the end of sentences. Placing \texttt{{REVISION}} at the beginning disrupts the natural flow of information, leading to confusion and reduced accuracy.

\paragraph{Comparative}
Additionally, we observe that DistilBERT is not
sensitive to comparatives.
In Table \ref{result_sentiment}, its accuracy
for
the two ``positive'' instantiations of the comparative test
type
are 0.67\% and 0.00\% 
whereas RoBERTa is close to 100\% accurate.

\paragraph{Specific Phrase}
Beyond these
grammatical features, we also find that specific words
or phrases can decrease accuracy.
For the 
\texttt{blown away by} template,
DistilBERT has accuracy
0\%, while both models are at 0\% for
\texttt{a perfect little atrocity}.
We suspect that the idiomaticity
of these expressions causes confusion. \texttt{blown away
  by}
is positive despite its negative literal meaning, and \texttt{a perfect
  little atrocity} is strongly negative despite containing positive words, probably misleading due to a combination of
words with contrasting terms.
This suggests that 
models struggle with phrases where literal
conflicts with conventional
meaning.

\paragraph{Irony}
Besides the patterns listed in the table,
models also frequently misinterpret
irony.
For instance,
DistilBERT misreads the sentence \texttt{Quite
  what the producers of this film were thinking when they
  chose to cast Nicholas Cage in the lead role is a
  mystery.} as positive, with a 69\% discrepancy in
positive sentiment score compared to the reference model.
However, due to the nuanced and context-dependent nature
 of irony, we were unable to formalize this into a specific test type.

\subsection{Toxicity Detection}
We select two \taskmodels for binary
(i.e., toxic vs.\ non-toxic)
toxicity detection.
We focus on
the ToxiGen datasets and models
\cite{hartvigsen-etal-2022-toxigen}
because of their diversity, which improves current
models' performance on toxicity detection. The first
\taskmodel is ToxDetect, which
\citet{hartvigsen-etal-2022-toxigen} further fine-tune on
ToxiGen train (the base is
a RoBERTa large model \cite{zhou-etal-2021-challenges}).
The second
is a much smaller DistilBERT model (also fine-tuned on
ToxiGen train). We test these models on 
ToxiGen test. The RoBERTa model achieves $77.6\%$ accuracy,
the DistilBERT model $67.7\%$. As a
reference model in the second step of \framework, we again
use 4-shot LLaMA2 7B; its accuracy is $83.0\%$.

Using \framework as we did for sentiment classification, we randomly sample 100,000 five-word
queries from ToxiGen train and generate sentences based
on them. Then we use 4-shot LLaMA2 7B and two \taskmodels, ToxDetect, and
finetuned DistilBERT, to classify these sentences for
toxicity.
Next, we compute the probability difference of the toxic label between the reference model and each \taskmodel separately, and focus on the top 10,000 sentences with the highest probability difference. 
Our annotators find that many
of the challenging toxic sentences are often aimed at
specific groups. In total, we identify 8 distinct patterns.

Table \ref{result_toxic} gives
\framework results for toxic language detection. We find that
DistilBERT 
performs worse than anticipated, being easily misled by almost
all tested templates. In contrast, RoBERTa
performs well for three ``Stereotypes and Directives'' test types except the template
\texttt{\{NATIONALITY\} should \{DENY HUMAN RIGHTS\}}.
When 
integrating variations such as
``Typos'' and ``Nonsense characters'', the models
display varying degrees of
confusion where
DistilBERT is more negatively affected than
RoBERTa.

\begin{table*}[]
\centering
\resizebox{\textwidth}{!}{%
\begin{tabular}{llrrc}
\toprule
\textbf{Test Type} & \multicolumn{1}{l}{\textbf{Template}} & \multicolumn{2}{c}{\textbf{Accuracy(\%)}} & \textbf{Gold} \\
\multicolumn{1}{l}{}&     & \textbf{DistilBERT}   & \textbf{RoBERTa}  & \textbf{Label}\\ \midrule
\multirow{4}{*}{\makecell{\textbf{Stereotypes and}\\ \textbf{Directives}}}
& $\left\{\begin{array}{l} \colorbox{lightmaroon}{\text{Asians}\textsubscript{\scalebox{0.5}{NATIONALITY}}} \\ \colorbox{lightkhaki}{\text{You}\textsubscript{\scalebox{0.5}{PERSONAL PRONOUN}}}\end{array}\right\}$ \{\colorbox{cantaloupe}{deserve to\textsubscript{\scalebox{0.5}{TERM OF ``DESERVE TO''}}}\} die. 
& $\begin{array}{r} \gradient{12.75} \\ \gradient{37.50}\end{array}$ & $\begin{array}{r} \gradient{100.00} \\ \gradient{100.00}\end{array}$ & $\begin{array}{r} \text{\emojibiohazard} \\ \text{\emojibiohazard}\end{array}$ \\
\gcmidrule(lr){2-2}
&  \{\colorbox{lightmaroon}{\text{Asians}\textsubscript{\scalebox{0.5}{NATIONALITY}}}\} should \{\colorbox{lightthistle}{be restricted from traveling freely\textsubscript{\scalebox{0.5}{DENY HUMAN RIHGHTS}}}\}.    & $\begin{array}{r} \gradient{24.51} \end{array}$ & $\begin{array}{r} \gradient{70.22} \end{array}$& $\begin{array}{r} \text{\emojibiohazard} \end{array}$ \\
\gcmidrule(lr){2-2}
&  \{\colorbox{lightmaroon}{\text{Asians}\textsubscript{\scalebox{0.5}{NATIONALITY}}}\} should not \{\colorbox{lightthistle}{be restricted from traveling freely\textsubscript{\scalebox{0.5}{DENY HUMAN RIHGHTS}}}\}. & $\begin{array}{r} \gradient{95.07} \end{array}$ & $\begin{array}{r} \gradient{97.50}\end{array}$ & $\begin{array}{r} \text{\emojipottedplant} \end{array}$\\ 
\midrule
\multirow{1}{*}{\textbf{Typos}} &  \{\colorbox{lightmaroon}{\text{Asians}\textsubscript{\scalebox{0.5}{NATIONALITY}}}\} \{\colorbox{petalpink}{deseve to\textsubscript{\scalebox{0.5}{``DESERVE'' TERM TYPO}}}\} die.& $\begin{array}{r} \gradient{0.70} \\ \end{array}$ & $\begin{array}{r} \gradient{92.75} \\ \end{array}$& $\begin{array}{r} \text{\emojibiohazard} \\ \end{array}$ \\
\midrule

\textbf{Reversal}& \{\colorbox{lightmaroon}{\text{Asians}\textsubscript{\scalebox{0.5}{NATIONALITY}}}\} deserve to die.\{\colorbox{watermelon}{That's wrong!\textsubscript{\scalebox{0.5}{DENY PHRASE}}}\}  & $\begin{array}{r}\gradient{88.07} \end{array}$&$\begin{array}{r} \gradient{11.36}  \end{array}$& $\begin{array}{r}\text{\emojipottedplant}\end{array}$ \\ 
\midrule
\multirow{2}{*}{\textbf{Nonsense Characters}}& \{\colorbox{lightmaroon}{\text{Asians}\textsubscript{\scalebox{0.5}{NATIONALITY}}}\} deserve to die.\{\colorbox{lightgold}{------>------------->-----/n\textsubscript{\scalebox{0.5}{STRANGE SUFFIX}}}\}& $\begin{array}{r} \gradient{0.00}  \end{array}$ & $\begin{array}{r} \gradient{5.85}  \end{array}$& $\begin{array}{r} \text{\emojibiohazard} \end{array}$ \\ & \{\colorbox{lightgold}{------>------------->-----/n\textsubscript{\scalebox{0.5}{STRANGE PREFIX}}}\}
\{\colorbox{lightmaroon}{\text{Asians}\textsubscript{\scalebox{0.5}{NATIONALITY}}}\} deserve to die.& $\begin{array}{r} \gradient{0.00}  \end{array}$ & $\begin{array}{r} \gradient{1.91}  \end{array}$& $\begin{array}{r} \text{\emojibiohazard} \end{array}$ \\
\midrule
\multirow{2}{*}{\textbf{Ethnic Slurs}} & \{\colorbox{lightsteelblue}{It seems this}{ \texttt{N-word}} \colorbox{lightsteelblue}{forgot his lunch today.\textsubscript{\scalebox{0.5}{NON-TOXIC SENTENCE}}}\}&$\begin{array}{r}\gradient{98.00}\end{array}$&$\begin{array}{r} \gradient{92.00} \end{array}$&  $\begin{array}{r} \text{\emojibiohazard
} \end{array}$ \\
& \{\colorbox{lightsteelblue}{It seems this} {Ching chong} \colorbox{lightsteelblue} {forgot his lunch today.\textsubscript{\scalebox{0.5}{NON-TOXIC SENTENCE}}}\}& $\begin{array}{r} \gradient{0.00} \end{array}$& $\begin{array}{r} \gradient{0.00} \end{array}$& $\begin{array}{r} \text{\emojibiohazard} \end{array}$ \\ 
\midrule
\multirow{2}{*}{\makecell{\textbf{Sentiment Analysis}\\ \textbf{Pattern}}}     & This \{\colorbox{lightgray}{book\textsubscript{\scalebox{0.50}{NOUN}}}\} is not  \{\colorbox{lightgreen}{nice\textsubscript{\scalebox{0.5}{POS ADJ}}}\}.   & $\begin{array}{r} \gradient{0.00}\end{array}$ & $\begin{array}{r} \gradient{99.25}\end{array}$ & $\begin{array}{r} \text{\emojipottedplant}\end{array}$ \\
\gcmidrule(lr){2-2}
& This \{\colorbox{lightgray}{book\textsubscript{\scalebox{0.50}{NOUN}}}\} is  \{\colorbox{lightpink}{awful\textsubscript{\scalebox{0.5}{NEG ADJ}}}\}.& $\begin{array}{r} \gradient{0.00}\end{array}$ & $\begin{array}{r} \gradient{8.99}  \end{array}$& $\begin{array}{r} \text{\emojipottedplant}\end{array}$ \\
\bottomrule
\end{tabular}%
}
\caption{\framework templates for the toxic language detection task and
	accuracy for two \taskmodels. Accuracy is percentage of sentences from a template that the model predicts correctly. \text{\emojibiohazard} represents the label ``toxic'', while \text{\emojipottedplant} represents the label ``non-toxic''.}

\label{result_toxic}
\end{table*}

\paragraph{Nonsense Characters}

We observed that during synthetic test generations in \framework, LLMs sometimes add random characters at the end of the generation. This led us to design tests adding nonsense characters to well-defined examples. When dealing with the ``Nonsense Characters'' test type, both \taskmodels perform poorly. 
We used random non-alphabetic strings of 60-70 characters as placeholders for the template, 
which may have diluted meaningful features, especially in short sentences, causing information to be overwhelmed. Both models achieved less than 6\% accuracy with this test type.
 In follow-up tests, we experimented with different numbers of nonsense characters. We found that inserting random non-alphabetic strings of length 10 at the beginning or end of a sentence instead of 60-70 had little effect on model performance, 
 with DistilBERT achieving 83.7\% accuracy and RoBERTa achieving 100\%. 
 RoBERTa was less affected by noise inserted at the end of the sentence. 

\begin{table}[]
\centering
\resizebox{\columnwidth}{!}{%
\begin{tabular}{lrr}
\toprule
\multicolumn{1}{c}{\multirow{2}{*}{\textbf{Ethnic Slurs}}} & \multicolumn{2}{c}{\textbf{Accuracy(\%)}} \\ \cline{2-3} 
\multicolumn{1}{c}{} & \multicolumn{1}{c}{\textbf{DistilBERT}} & \multicolumn{1}{c}{\textbf{RoBERTa}} \\ \midrule
White \texttt{N-word}/\texttt{N-word} wop & 100.00 & 100.00 \\
Ching chong & 0.00 & 0.00 \\
Ukro-Nazi & 48.00 & 88.00 \\
hillbilly & 0.00 & 2.00 \\ \bottomrule
\end{tabular}%
}
\caption{Part of the test results on the List of Ethnic Slurs indicates the model's failure to recognize certain offensive terms and its lack of sensitivity to contextually toxic language. For instance, while the term \texttt{``hillbilly''} may not seem offensive in its lexical form, it is widely recognized as a derogatory term.}
\label{tab:slurs_short}
\end{table}

\paragraph{Ethnic Slurs} Exploratory tests employing racial slurs, including  \textit{\texttt{N-word}} and \textit{Ching Chong}, yield mixed results. The latter proved more effective in deceiving the models, suggesting a potential vulnerability in the models' ability to recognize a broader spectrum of ethnic slurs.

We test words from Wikipedia's List of Ethnic
Slurs.\footnote{\href{https://en.wikipedia.org/wiki/List_of_ethnic_slurs}{https://en.wikipedia.org/wiki/List\_of\_ethnic\_slurs}}
Table~\ref{tab:slurs_short} gives a few examples,
see Table \ref{tab:slurs} in the
Appendix for full results.
Some of the phrases in the Wikipedia list may not qualify as insulting, but as can be seen, neither model performs well. 
This suggests a need for more diverse and comprehensive training datasets that include a wider array of ethnic slurs and culturally specific derogatory terms to enhance model sensitivity and accuracy.

\paragraph{Robustness to Typos} To evaluate robustness
against typos, we choose the sentence template
\texttt{\{NATIONALITY\} \{TERM OF "DESERVE TO"\} die.} as
RoBERTa performed well on it (100\%) and
randomly delete, insert and swap letters for key verbs to
mimic typos. Table \ref{result_toxic} shows that
typos can have a large impact on
DistilBERT and still a noticeable impact on RoBERTa
(a
drop to
92.75\%). This
indicates a lack of robustness.

\paragraph{Distinguishing Negative Sentiment from Toxicity} According to Table \ref{result_toxic}, in the test type Sentiment analysis pattern, sentences that are negative yet non-toxic are particularly effective in misleading the models. DistilBERT classifies sentences using negative sentiment expressions as toxic. RoBERTa fails for the template \texttt{This \{NOUN\} is \{NEG ADJ\}} (8.99\% accuracy). This phenomenon underscores a critical area for further refinement in distinguishing between negativity and toxicity.

\paragraph{Simile}
We also find rhetorical patterns in sentences that confuse
the model. In analyzing
\texttt{SynthTest\textsubscript{hard}} sentences for
toxicity detection, it is clear that the model struggles to understand similes. For instance, models cannot detect toxicity in sentence \texttt{a person with a disability is like the lockers in the ladies room in the front of the school, and they seem nice, but they are going to be awful because the doors are huge and difficult to operate with not enough space}. %
This demonstrates a shortfall in the model's ability to interpret and assess complex social contexts within similes, often leading to a failure to identify underlying racial or discriminatory undertones that can make such statements toxic.

\section{Conclusion}
We presented \framework, a semi-automatic framework that
utilizes the generative and generalization capabilities of
LLMs to decrease human workload and provide novel insights
into behavioral testing for identifying vulnerabilities in
classification models. Our experiments demonstrated that
despite achieving high accuracy on traditional benchmarks,
the models exhibit significant vulnerabilities related to
linguistic complexity and limited vocabulary. \framework has
proven effective across two diverse classification tasks, offering a valuable tool for model evaluation with reduced human effort.

\section{Limitations}
Although we tested \framework with two different classification tasks, both of them were binary classification tasks. For more complex and multi-categorization tasks, this framework would require some modifications. For example, if there are neutral labels in the sentiment analysis task, a more sophisticated method for identifying challenging subsets of the test set could be devised. This could involve using KL divergence as the discrepancy metric or utilizing multiple LLMs, similar to the self-consistency approach \citep{wangself}.

Besides, we actually found some important patterns such as simile and irony, but it is difficult to find templates for these patterns to batch-generate sentences, which leads to the lack of comprehensiveness of our experiment.

It is also worth mentioning that the identification of these patterns was not entirely automated. Some potential patterns may have been overlooked in the process, suggesting that a more comprehensive or refined approach might be necessary to capture the full spectrum of factors influencing sentence classification.

\section{Ethical Concerns}
This paper contains some toxic statements about specific groups, but these sentences were only used for testing and generated by the LLM. They are not the opinions of the authors. A warning has been placed prominently on the first page.

 \section*{Acknowledgments}
 This research was supported by the Deutsche Forschungsgemeinschaft DFG (grant SCHU 2246/14-1).
 
 \bibliographystyle{acl_natbib}
\bibliography{custom, anthology_1, anthology_2}

\begin{thebibliography}{53}
\providecommand{\natexlab}[1]{#1}

\bibitem[{Beizer(1995)}]{Beizer1995blackbox}
Boris Beizer. 1995.
\newblock \emph{Black-box testing - techniques for functional testing of
  software and systems}.
\newblock Wiley.

\bibitem[{Belinkov and Bisk(2018)}]{Belinkov2018noise}
Yonatan Belinkov and Yonatan Bisk. 2018.
\newblock \href {https://openreview.net/forum?id=BJ8vJebC-} {Synthetic and
  natural noise both break neural machine translation}.
\newblock In \emph{6th International Conference on Learning Representations,
  {ICLR} 2018, Vancouver, BC, Canada, April 30 - May 3, 2018, Conference Track
  Proceedings}. OpenReview.net.

\bibitem[{Brown et~al.(2020)Brown, Mann, Ryder, Subbiah, Kaplan, Dhariwal,
  Neelakantan, Shyam, Sastry, Askell, Agarwal, Herbert-Voss, Krueger, Henighan,
  Child, Ramesh, Ziegler, Wu, Winter, Hesse, Chen, Sigler, Litwin, Gray, Chess,
  Clark, Berner, McCandlish, Radford, Sutskever, and
  Amodei}]{NEURIPS2020_1457c0d6}
Tom Brown, Benjamin Mann, Nick Ryder, Melanie Subbiah, Jared~D Kaplan, Prafulla
  Dhariwal, Arvind Neelakantan, Pranav Shyam, Girish Sastry, Amanda Askell,
  Sandhini Agarwal, Ariel Herbert-Voss, Gretchen Krueger, Tom Henighan, Rewon
  Child, Aditya Ramesh, Daniel Ziegler, Jeffrey Wu, Clemens Winter, Chris
  Hesse, Mark Chen, Eric Sigler, Mateusz Litwin, Scott Gray, Benjamin Chess,
  Jack Clark, Christopher Berner, Sam McCandlish, Alec Radford, Ilya Sutskever,
  and Dario Amodei. 2020.
\newblock \href
  {https://proceedings.neurips.cc/paper_files/paper/2020/file/1457c0d6bfcb4967418bfb8ac142f64a-Paper.pdf}
  {Language models are few-shot learners}.
\newblock In \emph{Advances in Neural Information Processing Systems},
  volume~33, pages 1877--1901. Curran Associates, Inc.

\bibitem[{Clark et~al.(2021)Clark, August, Serrano, Haduong, Gururangan, and
  Smith}]{clark-etal-2021-thats}
Elizabeth Clark, Tal August, Sofia Serrano, Nikita Haduong, Suchin Gururangan,
  and Noah~A. Smith. 2021.
\newblock \href {https://doi.org/10.18653/v1/2021.acl-long.565} {All that{'}s
  {`}human{'} is not gold: Evaluating human evaluation of generated text}.
\newblock In \emph{Proceedings of the 59th Annual Meeting of the Association
  for Computational Linguistics and the 11th International Joint Conference on
  Natural Language Processing (Volume 1: Long Papers)}, pages 7282--7296,
  Online. Association for Computational Linguistics.

\bibitem[{Dang et~al.(2020)Dang, Moreno-Garc{\'\i}a, and De~la
  Prieta}]{dang2020sentiment}
Nhan~Cach Dang, Mar{\'\i}a~N Moreno-Garc{\'\i}a, and Fernando De~la Prieta.
  2020.
\newblock Sentiment analysis based on deep learning: A comparative study.
\newblock \emph{Electronics}, 9(3):483.

\bibitem[{D{\'\i}az and Hecht-Felella(2021)}]{diaz2021double}
{\'A}ngel D{\'\i}az and Laura Hecht-Felella. 2021.
\newblock Double standards in social media content moderation.
\newblock \emph{Brennan Center for Justice at New York University School of
  Law. https://www. brennancenter.
  org/our-work/research-reports/double-standards-socialmedia-content-moderation}.

\bibitem[{F{\"a}rber et~al.(2020)F{\"a}rber, Burkard, Jatowt, and
  Lim}]{farber2020multidimensional}
Michael F{\"a}rber, Victoria Burkard, Adam Jatowt, and Sora Lim. 2020.
\newblock A multidimensional dataset based on crowdsourcing for analyzing and
  detecting news bias.
\newblock In \emph{Proceedings of the 29th ACM international conference on
  information \& knowledge management}, pages 3007--3014.

\bibitem[{Ferrando et~al.(2023)Ferrando, Sperber, Setiawan, Telaar, and
  Hasan}]{ferrando-etal-2023-automating}
Javier Ferrando, Matthias Sperber, Hendra Setiawan, Dominic Telaar, and
  Sa{\v{s}}a Hasan. 2023.
\newblock \href {https://doi.org/10.18653/v1/2023.wmt-1.97} {Automating
  behavioral testing in machine translation}.
\newblock In \emph{Proceedings of the Eighth Conference on Machine
  Translation}, pages 1014--1030, Singapore. Association for Computational
  Linguistics.

\bibitem[{Founta et~al.(2018)Founta, Djouvas, Chatzakou, Leontiadis, Blackburn,
  Stringhini, Vakali, Sirivianos, and Kourtellis}]{founta2018large}
Antigoni Founta, Constantinos Djouvas, Despoina Chatzakou, Ilias Leontiadis,
  Jeremy Blackburn, Gianluca Stringhini, Athena Vakali, Michael Sirivianos, and
  Nicolas Kourtellis. 2018.
\newblock Large scale crowdsourcing and characterization of twitter abusive
  behavior.
\newblock In \emph{Proceedings of the international AAAI conference on web and
  social media}, volume~12.

\bibitem[{Gardner et~al.(2020)Gardner, Artzi, Basmov, Berant, Bogin, Chen,
  Dasigi, Dua, Elazar, Gottumukkala, Gupta, Hajishirzi, Ilharco, Khashabi, Lin,
  Liu, Liu, Mulcaire, Ning, Singh, Smith, Subramanian, Tsarfaty, Wallace,
  Zhang, and Zhou}]{gardner-etal-2020-evaluating}
Matt Gardner, Yoav Artzi, Victoria Basmov, Jonathan Berant, Ben Bogin, Sihao
  Chen, Pradeep Dasigi, Dheeru Dua, Yanai Elazar, Ananth Gottumukkala, Nitish
  Gupta, Hannaneh Hajishirzi, Gabriel Ilharco, Daniel Khashabi, Kevin Lin,
  Jiangming Liu, Nelson~F. Liu, Phoebe Mulcaire, Qiang Ning, Sameer Singh,
  Noah~A. Smith, Sanjay Subramanian, Reut Tsarfaty, Eric Wallace, Ally Zhang,
  and Ben Zhou. 2020.
\newblock \href {https://doi.org/10.18653/v1/2020.findings-emnlp.117}
  {Evaluating models{'} local decision boundaries via contrast sets}.
\newblock In \emph{Findings of the Association for Computational Linguistics:
  EMNLP 2020}, pages 1307--1323, Online. Association for Computational
  Linguistics.

\bibitem[{Hartmann et~al.(2023)Hartmann, Heitmann, Siebert, and
  Schamp}]{hartmann2023}
Jochen Hartmann, Mark Heitmann, Christian Siebert, and Christina Schamp. 2023.
\newblock \href {https://doi.org/10.1016/j.ijresmar.2022.05.005} {More than a
  feeling: Accuracy and application of sentiment analysis}.
\newblock \emph{International Journal of Research in Marketing}, 40(1):75--87.

\bibitem[{Hartvigsen et~al.(2022)Hartvigsen, Gabriel, Palangi, Sap, Ray, and
  Kamar}]{hartvigsen-etal-2022-toxigen}
Thomas Hartvigsen, Saadia Gabriel, Hamid Palangi, Maarten Sap, Dipankar Ray,
  and Ece Kamar. 2022.
\newblock \href {https://doi.org/10.18653/v1/2022.acl-long.234} {{T}oxi{G}en: A
  large-scale machine-generated dataset for adversarial and implicit hate
  speech detection}.
\newblock In \emph{Proceedings of the 60th Annual Meeting of the Association
  for Computational Linguistics (Volume 1: Long Papers)}, pages 3309--3326,
  Dublin, Ireland. Association for Computational Linguistics.

\bibitem[{He et~al.(2023)He, Ribeiro, and Khani}]{he-etal-2023-targeted}
Zexue He, Marco~Tulio Ribeiro, and Fereshte Khani. 2023.
\newblock \href {https://doi.org/10.18653/v1/2023.acl-long.474} {Targeted data
  generation: Finding and fixing model weaknesses}.
\newblock In \emph{Proceedings of the 61st Annual Meeting of the Association
  for Computational Linguistics (Volume 1: Long Papers)}, pages 8506--8520,
  Toronto, Canada. Association for Computational Linguistics.

\bibitem[{Heng et~al.(2024)Heng, Deng, Li, Yu, Li, Zhang, and
  Zhang}]{heng2024proggen}
Yuzhao Heng, Chunyuan Deng, Yitong Li, Yue Yu, Yinghao Li, Rongzhi Zhang, and
  Chao Zhang. 2024.
\newblock Proggen: Generating named entity recognition datasets step-by-step
  with self-reflexive large language models.
\newblock \emph{arXiv preprint arXiv:2403.11103}.

\bibitem[{Holtzman et~al.(2019)Holtzman, Buys, Du, Forbes, and
  Choi}]{holtzman2019curious}
Ari Holtzman, Jan Buys, Li~Du, Maxwell Forbes, and Yejin Choi. 2019.
\newblock The curious case of neural text degeneration.
\newblock In \emph{International Conference on Learning Representations}.

\bibitem[{Jawahar et~al.(2020)Jawahar, Abdul-Mageed, and
  Lakshmanan}]{jawahar-etal-2020-automatic}
Ganesh Jawahar, Muhammad Abdul-Mageed, and Laks Lakshmanan, V.S. 2020.
\newblock \href {https://doi.org/10.18653/v1/2020.coling-main.208} {Automatic
  detection of machine generated text: A critical survey}.
\newblock In \emph{Proceedings of the 28th International Conference on
  Computational Linguistics}, pages 2296--2309, Barcelona, Spain (Online).
  International Committee on Computational Linguistics.

\bibitem[{K et~al.(2022)K, Bhatt, Singh, Aditya, Dandapat, Sitaram, and
  Choudhury}]{k-etal-2022-multilingual}
Karthikeyan K, Shaily Bhatt, Pankaj Singh, Somak Aditya, Sandipan Dandapat,
  Sunayana Sitaram, and Monojit Choudhury. 2022.
\newblock \href {https://aclanthology.org/2022.findings-aacl.27} {Multilingual
  {C}heck{L}ist: Generation and evaluation}.
\newblock In \emph{Findings of the Association for Computational Linguistics:
  AACL-IJCNLP 2022}, pages 282--295, Online only. Association for Computational
  Linguistics.

\bibitem[{Kassner and Sch{\"u}tze(2020)}]{kassner-schutze-2020-negated}
Nora Kassner and Hinrich Sch{\"u}tze. 2020.
\newblock \href {https://doi.org/10.18653/v1/2020.acl-main.698} {Negated and
  misprimed probes for pretrained language models: Birds can talk, but cannot
  fly}.
\newblock In \emph{Proceedings of the 58th Annual Meeting of the Association
  for Computational Linguistics}, pages 7811--7818, Online. Association for
  Computational Linguistics.

\bibitem[{Kennedy et~al.(2018)Kennedy, Atari, Davani, Yeh, Omrani, Kim, Coombs,
  Havaldar, Portillo-Wightman, Gonzalez et~al.}]{kennedy2018gab}
Brendan Kennedy, Mohammad Atari, Aida~Mostafazadeh Davani, Leigh Yeh, Ali
  Omrani, Yehsong Kim, Kris Coombs, Shreya Havaldar, Gwenyth Portillo-Wightman,
  Elaine Gonzalez, et~al. 2018.
\newblock The gab hate corpus: A collection of 27k posts annotated for hate
  speech.
\newblock \emph{PsyArXiv. July}, 18.

\bibitem[{Kiela et~al.(2021)Kiela, Bartolo, Nie, Kaushik, Geiger, Wu, Vidgen,
  Prasad, Singh, Ringshia, Ma, Thrush, Riedel, Waseem, Stenetorp, Jia, Bansal,
  Potts, and Williams}]{kiela-etal-2021-dynabench}
Douwe Kiela, Max Bartolo, Yixin Nie, Divyansh Kaushik, Atticus Geiger,
  Zhengxuan Wu, Bertie Vidgen, Grusha Prasad, Amanpreet Singh, Pratik Ringshia,
  Zhiyi Ma, Tristan Thrush, Sebastian Riedel, Zeerak Waseem, Pontus Stenetorp,
  Robin Jia, Mohit Bansal, Christopher Potts, and Adina Williams. 2021.
\newblock \href {https://doi.org/10.18653/v1/2021.naacl-main.324} {Dynabench:
  Rethinking benchmarking in {NLP}}.
\newblock In \emph{Proceedings of the 2021 Conference of the North American
  Chapter of the Association for Computational Linguistics: Human Language
  Technologies}, pages 4110--4124, Online. Association for Computational
  Linguistics.

\bibitem[{K{\"o}ksal et~al.(2023)K{\"o}ksal, Aksitov, and
  Chang}]{koksal2023hallucination}
Abdullatif K{\"o}ksal, Renat Aksitov, and Chung-Ching Chang. 2023.
\newblock Hallucination augmented recitations for language models.
\newblock \emph{arXiv preprint arXiv:2311.07424}.

\bibitem[{Köksal et~al.(2024)Köksal, Schick, Korhonen, and
  Schütze}]{koksal2024longform}
Abdullatif Köksal, Timo Schick, Anna Korhonen, and Hinrich Schütze. 2024.
\newblock \href {https://arxiv.org/abs/2304.08460} {Longform: Effective
  instruction tuning with reverse instructions}.
\newblock \emph{Preprint}, arXiv:2304.08460.

\bibitem[{Lee et~al.(2024)Lee, Kim, Kim, Cho, and Kang}]{lee2024checkeval}
Yukyung Lee, Joonghoon Kim, Jaehee Kim, Hyowon Cho, and Pilsung Kang. 2024.
\newblock Checkeval: Robust evaluation framework using large language model via
  checklist.
\newblock \emph{arXiv preprint arXiv:2403.18771}.

\bibitem[{Li et~al.(2023)Li, Zhu, Lu, and Yin}]{li-etal-2023-synthetic}
Zhuoyan Li, Hangxiao Zhu, Zhuoran Lu, and Ming Yin. 2023.
\newblock \href {https://doi.org/10.18653/v1/2023.emnlp-main.647} {Synthetic
  data generation with large language models for text classification: Potential
  and limitations}.
\newblock In \emph{Proceedings of the 2023 Conference on Empirical Methods in
  Natural Language Processing}, pages 10443--10461, Singapore. Association for
  Computational Linguistics.

\bibitem[{Liu et~al.(2019)Liu, Ott, Goyal, Du, Joshi, Chen, Levy, Lewis,
  Zettlemoyer, and Stoyanov}]{liu2019roberta}
Yinhan Liu, Myle Ott, Naman Goyal, Jingfei Du, Mandar Joshi, Danqi Chen, Omer
  Levy, Mike Lewis, Luke Zettlemoyer, and Veselin Stoyanov. 2019.
\newblock Roberta: A robustly optimized bert pretraining approach.
\newblock \emph{arXiv preprint arXiv:1907.11692}.

\bibitem[{Maudslay and
  Cotterell(2021)}]{hall-maudslay-cotterell-2021-syntactic}
Rowan~Hall Maudslay and Ryan Cotterell. 2021.
\newblock \href {https://doi.org/10.18653/v1/2021.naacl-main.11} {Do syntactic
  probes probe syntax? experiments with jabberwocky probing}.
\newblock In \emph{Proceedings of the 2021 Conference of the North American
  Chapter of the Association for Computational Linguistics: Human Language
  Technologies}, pages 124--131, Online. Association for Computational
  Linguistics.

\bibitem[{Mu{\~n}oz-Ortiz et~al.(2023)Mu{\~n}oz-Ortiz,
  G{\'o}mez-Rodr{\'\i}guez, and Vilares}]{munoz2023contrasting}
Alberto Mu{\~n}oz-Ortiz, Carlos G{\'o}mez-Rodr{\'\i}guez, and David Vilares.
  2023.
\newblock Contrasting linguistic patterns in human and llm-generated text.
\newblock \emph{arXiv preprint arXiv:2308.09067}.

\bibitem[{Naveed et~al.(2023)Naveed, Khan, Qiu, Saqib, Anwar, Usman, Barnes,
  and Mian}]{naveed2023comprehensive}
Humza Naveed, Asad~Ullah Khan, Shi Qiu, Muhammad Saqib, Saeed Anwar, Muhammad
  Usman, Nick Barnes, and Ajmal Mian. 2023.
\newblock A comprehensive overview of large language models.
\newblock \emph{arXiv preprint arXiv:2307.06435}.

\bibitem[{Patterson et~al.(2021)Patterson, Gonzalez, Le, Liang, Munguia,
  Rothchild, So, Texier, and Dean}]{patterson2021carbon}
David Patterson, Joseph Gonzalez, Quoc Le, Chen Liang, Lluis-Miquel Munguia,
  Daniel Rothchild, David So, Maud Texier, and Jeff Dean. 2021.
\newblock Carbon emissions and large neural network training.
\newblock \emph{arXiv preprint arXiv:2104.10350}.

\bibitem[{Perez et~al.(2022)Perez, Huang, Song, Cai, Ring, Aslanides, Glaese,
  McAleese, and Irving}]{perez-etal-2022-red}
Ethan Perez, Saffron Huang, Francis Song, Trevor Cai, Roman Ring, John
  Aslanides, Amelia Glaese, Nat McAleese, and Geoffrey Irving. 2022.
\newblock \href {https://doi.org/10.18653/v1/2022.emnlp-main.225} {Red teaming
  language models with language models}.
\newblock In \emph{Proceedings of the 2022 Conference on Empirical Methods in
  Natural Language Processing}, pages 3419--3448, Abu Dhabi, United Arab
  Emirates. Association for Computational Linguistics.

\bibitem[{Rajpurkar et~al.(2018)Rajpurkar, Jia, and
  Liang}]{rajpurkar-etal-2018-know}
Pranav Rajpurkar, Robin Jia, and Percy Liang. 2018.
\newblock \href {https://doi.org/10.18653/v1/P18-2124} {Know what you don{'}t
  know: Unanswerable questions for {SQ}u{AD}}.
\newblock In \emph{Proceedings of the 56th Annual Meeting of the Association
  for Computational Linguistics (Volume 2: Short Papers)}, pages 784--789,
  Melbourne, Australia. Association for Computational Linguistics.

\bibitem[{Ribeiro et~al.(2019)Ribeiro, Guestrin, and
  Singh}]{ribeiro-etal-2019-red}
Marco~Tulio Ribeiro, Carlos Guestrin, and Sameer Singh. 2019.
\newblock \href {https://doi.org/10.18653/v1/P19-1621} {Are red roses red?
  evaluating consistency of question-answering models}.
\newblock In \emph{Proceedings of the 57th Annual Meeting of the Association
  for Computational Linguistics}, pages 6174--6184, Florence, Italy.
  Association for Computational Linguistics.

\bibitem[{Ribeiro and Lundberg(2022)}]{ribeiro-lundberg-2022-adaptive}
Marco~Tulio Ribeiro and Scott Lundberg. 2022.
\newblock \href {https://doi.org/10.18653/v1/2022.acl-long.230} {Adaptive
  testing and debugging of {NLP} models}.
\newblock In \emph{Proceedings of the 60th Annual Meeting of the Association
  for Computational Linguistics (Volume 1: Long Papers)}, pages 3253--3267,
  Dublin, Ireland. Association for Computational Linguistics.

\bibitem[{Ribeiro et~al.(2020)Ribeiro, Wu, Guestrin, and
  Singh}]{ribeiro-etal-2020-beyond}
Marco~Tulio Ribeiro, Tongshuang Wu, Carlos Guestrin, and Sameer Singh. 2020.
\newblock \href {https://doi.org/10.18653/v1/2020.acl-main.442} {Beyond
  accuracy: Behavioral testing of {NLP} models with {C}heck{L}ist}.
\newblock In \emph{Proceedings of the 58th Annual Meeting of the Association
  for Computational Linguistics}, pages 4902--4912, Online. Association for
  Computational Linguistics.

\bibitem[{Roelofs(2019)}]{roelofs2019measuring}
Rebecca Roelofs. 2019.
\newblock \emph{Measuring Generalization and overfitting in Machine learning}.
\newblock University of California, Berkeley.

\bibitem[{Rogers et~al.(2020)Rogers, Kovaleva, and
  Rumshisky}]{rogers-etal-2020-primer}
Anna Rogers, Olga Kovaleva, and Anna Rumshisky. 2020.
\newblock \href {https://doi.org/10.1162/tacl_a_00349} {A primer in
  {BERT}ology: What we know about how {BERT} works}.
\newblock \emph{Transactions of the Association for Computational Linguistics},
  8:842--866.

\bibitem[{R{\"o}ttger et~al.(2021)R{\"o}ttger, Vidgen, Nguyen, Waseem,
  Margetts, and Pierrehumbert}]{rottger-etal-2021-hatecheck}
Paul R{\"o}ttger, Bertie Vidgen, Dong Nguyen, Zeerak Waseem, Helen Margetts,
  and Janet Pierrehumbert. 2021.
\newblock \href {https://doi.org/10.18653/v1/2021.acl-long.4} {{H}ate{C}heck:
  Functional tests for hate speech detection models}.
\newblock In \emph{Proceedings of the 59th Annual Meeting of the Association
  for Computational Linguistics and the 11th International Joint Conference on
  Natural Language Processing (Volume 1: Long Papers)}, pages 41--58, Online.
  Association for Computational Linguistics.

\bibitem[{Rudinger et~al.(2017)Rudinger, May, and
  Van~Durme}]{rudinger-etal-2017-social}
Rachel Rudinger, Chandler May, and Benjamin Van~Durme. 2017.
\newblock \href {https://doi.org/10.18653/v1/W17-1609} {Social bias in elicited
  natural language inferences}.
\newblock In \emph{Proceedings of the First {ACL} Workshop on Ethics in Natural
  Language Processing}, pages 74--79, Valencia, Spain. Association for
  Computational Linguistics.

\bibitem[{Sanh et~al.(2019)Sanh, Debut, Chaumond, and
  Wolf}]{sanh2019distilbert}
Victor Sanh, Lysandre Debut, Julien Chaumond, and Thomas Wolf. 2019.
\newblock Distilbert, a distilled version of bert: smaller, faster, cheaper and
  lighter.
\newblock \emph{arXiv preprint arXiv:1910.01108}.

\bibitem[{Sap et~al.(2019)Sap, Card, Gabriel, Choi, and
  Smith}]{sap-etal-2019-risk}
Maarten Sap, Dallas Card, Saadia Gabriel, Yejin Choi, and Noah~A. Smith. 2019.
\newblock \href {https://doi.org/10.18653/v1/P19-1163} {The risk of racial bias
  in hate speech detection}.
\newblock In \emph{Proceedings of the 57th Annual Meeting of the Association
  for Computational Linguistics}, pages 1668--1678, Florence, Italy.
  Association for Computational Linguistics.

\bibitem[{Socher et~al.(2013)Socher, Perelygin, Wu, Chuang, Manning, Ng, and
  Potts}]{socher-etal-2013-recursive}
Richard Socher, Alex Perelygin, Jean Wu, Jason Chuang, Christopher~D. Manning,
  Andrew Ng, and Christopher Potts. 2013.
\newblock \href {https://aclanthology.org/D13-1170} {Recursive deep models for
  semantic compositionality over a sentiment treebank}.
\newblock In \emph{Proceedings of the 2013 Conference on Empirical Methods in
  Natural Language Processing}, pages 1631--1642, Seattle, Washington, USA.
  Association for Computational Linguistics.

\bibitem[{Strubell et~al.(2020)Strubell, Ganesh, and
  McCallum}]{strubell2020energy}
Emma Strubell, Ananya Ganesh, and Andrew McCallum. 2020.
\newblock Energy and policy considerations for modern deep learning research.
\newblock In \emph{Proceedings of the AAAI conference on artificial
  intelligence}, volume~34, pages 13693--13696.

\bibitem[{Torralba and Efros(2011)}]{Torralba2011dataset}
Antonio Torralba and Alexei~A. Efros. 2011.
\newblock \href {https://doi.org/10.1109/CVPR.2011.5995347} {Unbiased look at
  dataset bias}.
\newblock In \emph{The 24th {IEEE} Conference on Computer Vision and Pattern
  Recognition, {CVPR} 2011, Colorado Springs, CO, USA, 20-25 June 2011}, pages
  1521--1528. {IEEE} Computer Society.

\bibitem[{Touvron et~al.(2023)Touvron, Martin, Stone, Albert, Almahairi,
  Babaei, Bashlykov, Batra, Bhargava, Bhosale et~al.}]{touvron2023llama}
Hugo Touvron, Louis Martin, Kevin Stone, Peter Albert, Amjad Almahairi, Yasmine
  Babaei, Nikolay Bashlykov, Soumya Batra, Prajjwal Bhargava, Shruti Bhosale,
  et~al. 2023.
\newblock Llama 2: Open foundation and fine-tuned chat models.
\newblock \emph{arXiv preprint arXiv:2307.09288}.

\bibitem[{Trinh et~al.(2024)Trinh, Wu, Le, He, and Luong}]{Trinh2024}
Trieu~H. Trinh, Yuhuai Wu, Quoc~V. Le, He~He, and Thang Luong. 2024.
\newblock \href {https://doi.org/10.1038/s41586-023-06747-5} {Solving olympiad
  geometry without human demonstrations}.
\newblock \emph{Nature}, 625(7995):476--482.

\bibitem[{Wang et~al.(2023)Wang, Wei, Schuurmans, Le, Chi, Narang, Chowdhery,
  and Zhou}]{wangself}
Xuezhi Wang, Jason Wei, Dale Schuurmans, Quoc~V Le, Ed~H. Chi, Sharan Narang,
  Aakanksha Chowdhery, and Denny Zhou. 2023.
\newblock \href {https://openreview.net/forum?id=1PL1NIMMrw} {Self-consistency
  improves chain of thought reasoning in language models}.
\newblock In \emph{The Eleventh International Conference on Learning
  Representations}.

\bibitem[{Weissweiler et~al.(2024)Weissweiler, Köksal, and
  Schütze}]{weissweiler2024hybrid}
Leonie Weissweiler, Abdullatif Köksal, and Hinrich Schütze. 2024.
\newblock \href {https://arxiv.org/abs/2403.06965} {Hybrid human-llm corpus
  construction and llm evaluation for rare linguistic phenomena}.
\newblock \emph{Preprint}, arXiv:2403.06965.

\bibitem[{Whitehouse et~al.(2023)Whitehouse, Choudhury, and
  Aji}]{whitehouse-etal-2023-llm}
Chenxi Whitehouse, Monojit Choudhury, and Alham Aji. 2023.
\newblock \href {https://doi.org/10.18653/v1/2023.emnlp-main.44} {{LLM}-powered
  data augmentation for enhanced cross-lingual performance}.
\newblock In \emph{Proceedings of the 2023 Conference on Empirical Methods in
  Natural Language Processing}, pages 671--686, Singapore. Association for
  Computational Linguistics.

\bibitem[{Wu et~al.(2019)Wu, Ribeiro, Heer, and Weld}]{wu-etal-2019-errudite}
Tongshuang Wu, Marco~Tulio Ribeiro, Jeffrey Heer, and Daniel Weld. 2019.
\newblock \href {https://doi.org/10.18653/v1/P19-1073} {{E}rrudite: Scalable,
  reproducible, and testable error analysis}.
\newblock In \emph{Proceedings of the 57th Annual Meeting of the Association
  for Computational Linguistics}, pages 747--763, Florence, Italy. Association
  for Computational Linguistics.

\bibitem[{Yang et~al.(2022)Yang, Haque, Song, Yang, and
  Liu}]{yang-etal-2022-testaug}
Guanqun Yang, Mirazul Haque, Qiaochu Song, Wei Yang, and Xueqing Liu. 2022.
\newblock \href {https://aclanthology.org/2022.coling-1.307} {{T}est{A}ug: A
  framework for augmenting capability-based {NLP} tests}.
\newblock In \emph{Proceedings of the 29th International Conference on
  Computational Linguistics}, pages 3480--3495, Gyeongju, Republic of Korea.
  International Committee on Computational Linguistics.

\bibitem[{Ye et~al.(2022)Ye, Gao, Li, Xu, Feng, Wu, Yu, and
  Kong}]{ye-etal-2022-zerogen}
Jiacheng Ye, Jiahui Gao, Qintong Li, Hang Xu, Jiangtao Feng, Zhiyong Wu, Tao
  Yu, and Lingpeng Kong. 2022.
\newblock \href {https://doi.org/10.18653/v1/2022.emnlp-main.801} {{Z}ero{G}en:
  Efficient zero-shot learning via dataset generation}.
\newblock In \emph{Proceedings of the 2022 Conference on Empirical Methods in
  Natural Language Processing}, pages 11653--11669, Abu Dhabi, United Arab
  Emirates. Association for Computational Linguistics.

\bibitem[{Yu et~al.(2023)Yu, Zhuang, Zhang, Meng, Ratner, Krishna, Shen, and
  Zhang}]{yu2023llmgen}
Yue Yu, Yuchen Zhuang, Jieyu Zhang, Yu~Meng, Alexander~J. Ratner, Ranjay
  Krishna, Jiaming Shen, and Chao Zhang. 2023.
\newblock \href
  {http://papers.nips.cc/paper\_files/paper/2023/hash/ae9500c4f5607caf2eff033c67daa9d7-Abstract-Datasets\_and\_Benchmarks.html}
  {Large language model as attributed training data generator: {A} tale of
  diversity and bias}.
\newblock In \emph{Advances in Neural Information Processing Systems 36: Annual
  Conference on Neural Information Processing Systems 2023, NeurIPS 2023, New
  Orleans, LA, USA, December 10 - 16, 2023}.

\bibitem[{Zhou et~al.(2021)Zhou, Sap, Swayamdipta, Choi, and
  Smith}]{zhou-etal-2021-challenges}
Xuhui Zhou, Maarten Sap, Swabha Swayamdipta, Yejin Choi, and Noah Smith. 2021.
\newblock \href {https://doi.org/10.18653/v1/2021.eacl-main.274} {Challenges in
  automated debiasing for toxic language detection}.
\newblock In \emph{Proceedings of the 16th Conference of the European Chapter
  of the Association for Computational Linguistics: Main Volume}, pages
  3143--3155, Online. Association for Computational Linguistics.

\end{thebibliography}

\clearpage
\appendix

\section{Appendix}
\label{sec:prompt}

\subsection{Prompts Design}

\textbf{Prompt for sentiment classification task:}

Question: Find the sentiment of this text. Answer with positive or negative: that is far too tragic to merit such superficial treatment

Answer: negative

Question: Find the sentiment of this text. Answer with positive or negative: a smile on your face

Answer: positive

Question: Find the sentiment of this text. Answer with positive or negative: saw how bad this movie was

Answer: negative

Question: Find the sentiment of this text. Answer with positive or negative: the greatest musicians

Answer: positive

Question: Find the sentiment of this text. Answer with positive or negative: {text}

Answer:
\\

\textbf{Prompt for toxic language detection task:}

Determine if the following text includes hate speech or offensive language. Respond with 'Yes' if it contains such language, or 'No' if it is free from hate speech and offensive language.

Text: ``japanese or korean people are fucking weird''

Answer: Yes

Text: ``the most dangerous job in america is fishing''

Answer: No

Text: ``if a girl is raped, it is her fault because she was wearing a short skirt and a low cut blouse and she was walking alone at night''

Answer: Yes

Text: ``i'll get to that at the women 's meeting tomorrow''

Answer: No

\subsection{Accuracy of Adjectives}

We measure the frequency of specific adjectives in sentences where the RoBERTa-Large-based model failed across all test types and find that negative adjectives overall have lower accuracy compared to positive adjectives. The results is shown in Table \ref{tab: confusion_ratio_pos_complete}.
\begin{table}[t]
\centering
\resizebox{0.6\columnwidth}{!}{%
\begin{tabular}{lll}
\toprule
 & \textbf{ADJ} & \textbf{Accuracy(\%)} \\ \midrule
\multirow{36}{*}{\textbf{POS ADJ}} & app-ealing & 70.2309 \\
 & inv-iting & 65.2861 \\
 & f-avorable & 59.8896 \\
 & ide-al & 57.1285 \\
 & joy-ful & 56.0241 \\
 & en-chant-ing & 53.4137 \\
 & ch-arming & 49.0462 \\
 & ex-h-ilar-ating & 48.8454 \\
 & super & 48.5693 \\
 & imp-ressive & 47.7661 \\
 & super-ior & 46.2098 \\
 & pleasant & 46.1345 \\
 & ad-mir-able & 44.5532 \\
 & perfect & 42.5201 \\
 & g-orge-ous & 41.0392 \\
 & re-fres-hing & 40.0602 \\
 & great & 39.1315 \\
 & ple-asing & 39.0311 \\
 & am-azing & 36.8725 \\
 & br-ill-iant & 36.8223 \\
 & aw-esome & 36.1446 \\
 & w-onder-ful & 35.9689 \\
 & s-atisf-ying & 35.8685 \\
 & inc-redible & 35.3916 \\
 & del-ight-ful & 34.9398 \\
 & extra-ordinary & 34.739 \\
 & fab-ulous & 34.5131 \\
 & rem-arkable & 34.4378 \\
 & mar-vel-ous & 34.4127 \\
 & ex-cellent & 33.2329 \\
 & f-ant-astic & 32.8062 \\
 & except-ional & 31.3002 \\
 & super-b & 31.0492 \\
 & ter-r-ific & 30.9488 \\
 & ex-empl-ary & 30.246 \\
 & out-standing & 29.4177 \\ \midrule
\multirow{17}{*}{\textbf{NEG ADJ}} & bor-ing & 29.8008 \\
 & du-ll & 29.0386 \\
 & offensive & 21.6622 \\
 & medi-ocre & 20.9491 \\
 & sh-oddy & 19.7443 \\
 & dis-agree-able & 17.5805 \\
 & dis-app-oint-ing & 17.2363 \\
 & un-exc-iting & 16.9904 \\
 & ter-rible & 15.4659 \\
 & un-imp-ressive & 15.3922 \\
 & aw-ful & 15.1217 \\
 & d-read-ful & 14.3349 \\
 & un-interesting & 14.1628 \\
 & app-alling & 13.3514 \\
 & un-acceptable & 13.2776 \\
 & hor-rend-ous & 12.4908 \\
 & pat-hetic & 11.2368 \\ \bottomrule
\end{tabular}%
}
\caption{Accuracy of positive adjectives in sentiment analysis task for RoBERTa-large-based model.}
\label{tab: confusion_ratio_pos_complete}
\end{table}

\subsection{Ethnic Slurs}

We conducted a more comprehensive test of Wikipedia List of Ethnic Slurs, where we use 60 non-toxic sentences as templates to generate sentences to evaluate the performance of the models. The results in Table \ref{tab:slurs} show that both models are not sensitive enough to many ethnic slurs. Of course, some of the words in the list may have other non-toxic meanings, for example, \texttt{``Apple''} can also mean food, and \texttt{``Ann''} may really be a person's name.

The accuracy \texttt{2.00\%}  appears frequently, because the single pattern \texttt{``\{ETHNIC SLURS\}'s dog is barking.''} is often incorrectly determined as toxic.

\vfill\null

\tiny
\tablefirsthead{%
    \toprule
    \textbf{Ethnic Slurs} & \multicolumn{2}{c}
    {\textbf{Accuracy(\%)}} \\
    \cline{2-3}
     & \textbf{Distilbert} & \textbf{RoBERTa} \\
    \midrule
}
\tablehead{%
    \multicolumn{3}{l}{\small\sl continued from previous page} \\
    \toprule
    \textbf{Ethnic Slurs} & \multicolumn{2}{c}
    {\textbf{Accuracy(\%)}} \\
    \cline{2-3}
     & \textbf{Distilbert} & \textbf{RoBERTa} \\
    \midrule
}
\tabletail{%
\bottomrule
\multicolumn{3}{r}{\small\sl continued on next page} \\
}
\tablelasttail{\bottomrule}
\begin{supertabular}{lrr}
Abbie & 0.00 & 0.00 \\
ABC & 0.00 & 2.00 \\
ABCD & 0.00 & 2.00 \\
Abid/Abeed & 0.00 & 0.00 \\
Abo/Abbo & 0.00 & 0.00 \\
Afro engineering & 56.00 & 50.00 \\
Ah Chah & 0.00 & 0.00 \\
Ali Baba & 0.00 & 0.00 \\
Alligator bait & 2.00 & 4.00 \\
Alpine Serb & 2.00 & 2.00 \\
AmaLawu & 2.00 & 0.00 \\
Ang mo & 0.00 & 0.00 \\
Ann & 0.00 & 0.00 \\
Annamite, mites & 4.00 & 0.00 \\
Ape & 6.00 & 2.00 \\
Apple & 0.00 & 0.00 \\
Arabush & 2.00 & 0.00 \\
Argie & 2.00 & 0.00 \\
Armo & 0.00 & 0.00 \\
Asing, Aseng & 0.00 & 0.00 \\
Nazi & 28.00 & 92.00 \\
Aunt Jemima & 0.00 & 0.00 \\
Bachicha & 0.00 & 0.00 \\
Baiano & 0.00 & 0.00 \\
Bamboula & 0.00 & 0.00 \\
Banaan & 0.00 & 0.00 \\
Balija & 0.00 & 0.00 \\
Banana & 0.00 & 0.00 \\
Banderite & 0.00 & 2.00 \\
Barbarian & 14.00 & 2.00 \\
Beaner & 0.00 & 0.00 \\
Bimbo & 2.00 & 74.00 \\
Black buck & 8.00 & 0.00 \\
Bluegum & 0.00 & 4.00 \\
Boche & 0.00 & 2.00 \\
Boeotian & 0.00 & 0.00 \\
Boerehater & 0.00 & 0.00 \\
Bog & 6.00 & 0.00 \\
Bogate & 2.00 & 0.00 \\
Bohunk & 0.00 & 0.00 \\
Bong & 0.00 & 0.00 \\
Boong & 2.00 & 0.00 \\
Boonga & 0.00 & 0.00 \\
Bootlip & 2.00 & 0.00 \\
Bougnoule & 0.00 & 0.00 \\
Bounty bar & 2.00 & 0.00 \\
Bozgor & 0.00 & 0.00 \\
Brownie & 0.00 & 0.00 \\
Buckwheat & 2.00 & 0.00 \\
Buddhahead & 2.00 & 0.00 \\
Buckra & 0.00 & 0.00 \\
Bulbash & 0.00 & 0.00 \\
Bule & 0.00 & 0.00 \\
Bumbay & 6.00 & 0.00 \\
Burrhead & 2.00 & 0.00 \\
Bushy & 2.00 & 0.00 \\
Cabbage Eater & 4.00 & 0.00 \\
Canaca & 0.00 & 0.00 \\
Camel jockey & 2.00 & 0.00 \\
Carcamano & 0.00 & 0.00 \\
Chankoro & 0.00 & 0.00 \\
Charlie & 0.00 & 0.00 \\
China Swede & 2.00 & 2.00 \\
Chee-chee & 0.00 & 0.00 \\
Cheese-eating surrender monkeys & 58.00 & 4.00 \\
Chefur (čefur) & 0.00 & 0.00 \\
Tsekwa / Chekwa & 0.00 & 0.00 \\
Chernozhopy & 0.00 & 0.00 \\
Chilote & 0.00 & 0.00 \\
Chinaman & 6.00 & 2.00 \\
Ching chong & 0.00 & 0.00 \\
Chink & 50.00 & 0.00 \\
Chinky & 58.00 & 0.00 \\
Chonky & 24.00 & 0.00 \\
Christ-killer & 2.00 & 30.00 \\
Choc-ice & 0.00 & 0.00 \\
Cholo & 0.00 & 0.00 \\
Chon & 0.00 & 0.00 \\
Chow & 6.00 & 0.00 \\
Chuchmek & 0.00 & 0.00 \\
Chug & 0.00 & 0.00 \\
Chukhna & 0.00 & 0.00 \\
Churka & 0.00 & 0.00 \\
Ciapaty, ciapak & 0.00 & 0.00 \\
Cigányforma & 0.00 & 0.00 \\
Cigány népek & 0.00 & 0.00 \\
Cioară & 0.00 & 0.00 \\
Cina & 0.00 & 0.00 \\
Coconut & 0.00 & 0.00 \\
Pacific Islander & 4.00 & 2.00 \\
Coño & 0.00 & 0.00 \\
Coolie & 0.00 & 0.00 \\
Coon & 0.00 & 0.00 \\
Coonass & 0.00 & 0.00 \\
Coreano & 0.00 & 0.00 \\
Cotton picker & 2.00 & 2.00 \\
Cracker & 4.00 & 0.00 \\
Crow & 2.00 & 0.00 \\
Crucco & 0.00 & 0.00 \\
Culchie & 0.00 & 0.00 \\
Curepí & 0.00 & 0.00 \\
Curry-muncher & 2.00 & 2.00 \\
Cushi & 0.00 & 0.00 \\
Czarnuch & 0.00 & 0.00 \\
Dago & 0.00 & 0.00 \\
Dal Khor & 0.00 & 0.00 \\
Dalle, Batak Dalle & 0.00 & 0.00 \\
darky & 0.00 & 0.00 \\
Dhoti & 0.00 & 0.00 \\
Dink & 0.00 & 2.00 \\
Dogan, dogun & 2.00 & 0.00 \\
Dothead, Dot & 0.00 & 0.00 \\
Dune coon & 4.00 & 0.00 \\
Eight ball & 0.00 & 0.00 \\
Engelsman & 0.00 & 0.00 \\
Eyetie & 0.00 & 0.00 \\
Fankui & 0.00 & 0.00 \\
Farang & 0.00 & 0.00 \\
Fenian & 8.00 & 0.00 \\
Festival children & 0.00 & 2.00 \\
verlan & 0.00 & 0.00 \\
Fjellabe & 0.00 & 0.00 \\
Flip & 0.00 & 0.00 \\
Franchute & 0.00 & 0.00 \\
Frenk & 0.00 & 2.00 \\
Fritz & 0.00 & 0.00 \\
Frog & 0.00 & 0.00 \\
Fuzzy-Wuzzy & 0.00 & 0.00 \\
Gabacho & 0.00 & 0.00 \\
Gabel & 0.00 & 0.00 \\
Gadjo & 0.00 & 0.00 \\
Gaijin & 0.00 & 4.00 \\
Galla & 0.00 & 0.00 \\
Gam, Gammat & 0.00 & 0.00 \\
Gans & 0.00 & 0.00 \\
Garoi & 0.00 & 0.00 \\
Geomdung-i & 0.00 & 0.00 \\
Gexhë & 0.00 & 0.00 \\
Gin & 0.00 & 0.00 \\
Gin jockey & 2.00 & 2.00 \\
Godon & 0.00 & 0.00 \\
Golliwog & 0.00 & 0.00 \\
Gook & 94.00 & 0.00 \\
Goombah & 4.00 & 0.00 \\
Gora & 0.00 & 0.00 \\
Goy & 0.00 & 0.00 \\
Grago & 6.00 & 0.00 \\
Greaser & 12.00 & 10.00 \\
Greenhorn & 0.00 & 0.00 \\
Gringo & 0.00 & 8.00 \\
Groid & 0.00 & 2.00 \\
Gub, Gubba & 2.00 & 0.00 \\
Guizi & 0.00 & 0.00 \\
Guido & 0.00 & 0.00 \\
Guinea & 2.00 & 0.00 \\
Gummihals & 2.00 & 0.00 \\
Gusano & 0.00 & 0.00 \\
Gweilo & 0.00 & 0.00 \\
Gwer & 0.00 & 0.00 \\
Gyp/Gip & 2.00 & 0.00 \\
Gyopo, Kyopo & 0.00 & 0.00 \\
Gypsy & 0.00 & 0.00 \\
Hairyback & 52.00 & 0.00 \\
Hajji & 0.00 & 0.00 \\
Half-breed & 16.00 & 4.00 \\
Half-caste & 20.00 & 0.00 \\
Haole & 2.00 & 2.00 \\
Heeb, Hebe & 0.00 & 0.00 \\
Heigui & 0.00 & 0.00 \\
Heukhyeong & 0.00 & 0.00 \\
Hevosmies & 0.00 & 0.00 \\
Hike & 0.00 & 0.00 \\
Hillbilly & 0.00 & 2.00 \\
Honky & 0.00 & 0.00 \\
Hori & 0.00 & 0.00 \\
Hottentot, Hotnot & 2.00 & 0.00 \\
Houtkop & 0.00 & 0.00 \\
Huan-a, Huana & 0.00 & 0.00 \\
Huinca & 0.00 & 0.00 \\
Hujaa & 0.00 & 0.00 \\
Hun & 0.00 & 0.00 \\
Hunky & 16.00 & 0.00 \\
Hymie & 0.00 & 0.00 \\
Ikey & 4.00 & 0.00 \\
Ikey-mo & 0.00 & 0.00 \\
Indon & 0.00 & 0.00 \\
Indognesial & 0.00 & 2.00 \\
Intsik & 0.00 & 0.00 \\
Inyenzi & 0.00 & 0.00 \\
Injun & 0.00 & 0.00 \\
Itaker & 0.00 & 0.00 \\
Jackeen & 0.00 & 0.00 \\
Jakun & 0.00 & 0.00 \\
Jamet & 0.00 & 0.00 \\
Japa & 0.00 & 0.00 \\
Jap & 12.00 & 0.00 \\
Japie & 0.00 & 0.00 \\
Jareer & 0.00 & 0.00 \\
Jerry & 0.00 & 0.00 \\
Jewboy & 14.00 & 4.00 \\
Jidan & 0.00 & 0.00 \\
Jigaboo & 2.00 & 2.00 \\
Jim Crow & 6.00 & 2.00 \\
Jjangkkae & 0.00 & 0.00 \\
Jjokbari & 0.00 & 0.00 \\
Jock & 0.00 & 0.00 \\
Jungle bunny & 0.00 & 0.00 \\
Jutku & 2.00 & 0.00 \\
Kaew & 0.00 & 0.00 \\
Kaffir & 0.00 & 0.00 \\
Kaffir boetie & 0.00 & 0.00 \\
Kalar & 0.00 & 0.00 \\
Kalia & 0.00 & 0.00 \\
Katwa & 0.00 & 0.00 \\
Kanaka & 0.00 & 0.00 \\
Kanake & 0.00 & 0.00 \\
Kano & 0.00 & 0.00 \\
Kaouiche & 0.00 & 0.00 \\
Käskopp & 0.00 & 0.00 \\
Katsap & 0.00 & 0.00 \\
Kebab & 0.00 & 0.00 \\
Keko & 0.00 & 0.00 \\
Keling & 0.00 & 0.00 \\
Kemosabe & 0.00 & 0.00 \\
Kettō & 0.00 & 0.00 \\
Russian & 0.00 & 4.00 \\
Kharkhuwa & 0.00 & 0.00 \\
Khokhol & 0.00 & 0.00 \\
Ikula & 0.00 & 0.00 \\
Kike & 0.00 & 0.00 \\
Kimchi & 0.00 & 0.00 \\
Kıro & 0.00 & 0.00 \\
Knacker & 2.00 & 2.00 \\
Kojaengi & 0.00 & 0.00 \\
Kolorad & 0.00 & 0.00 \\
Krankie & 0.00 & 2.00 \\
Krakkemut & 0.00 & 0.00 \\
Kraut & 0.00 & 0.00 \\
Kuronbō & 0.00 & 0.00 \\
Kkamdungi & 0.00 & 0.00 \\
Labus & 0.00 & 0.00 \\
Laowai & 0.00 & 0.00 \\
Land thief & 34.00 & 4.00 \\
Lapp & 0.00 & 0.00 \\
Lebo, Leb & 2.00 & 0.00 \\
Leupe lonko & 2.00 & 0.00 \\
Limey & 0.00 & 0.00 \\
Locust & 6.00 & 0.00 \\
Londo & 0.00 & 0.00 \\
Lubra & 0.00 & 0.00 \\
Lundy & 0.00 & 0.00 \\
Lugan & 0.00 & 0.00 \\
Mabuno/Mahbuno & 0.00 & 0.00 \\
Macaca & 0.00 & 0.00 \\
Macaronar & 0.00 & 0.00 \\
Majus & 0.00 & 0.00 \\
Malakh-khor & 0.00 & 0.00 \\
Malau & 0.00 & 0.00 \\
Malaun & 0.00 & 0.00 \\
Malingsia & 0.00 & 0.00 \\
Malon & 0.00 & 0.00 \\
Mangal & 0.00 & 0.00 \\
Manne & 0.00 & 0.00 \\
Marokaki & 2.00 & 0.00 \\
Maruta & 0.00 & 0.00 \\
Mau-Mau & 0.00 & 0.00 \\
Mayate/Mayatero & 0.00 & 0.00 \\
Mayonnaise Monkey & 2.00 & 0.00 \\
Mick & 0.00 & 0.00 \\
Mocro & 2.00 & 0.00 \\
Mof & 2.00 & 0.00 \\
Momo & 0.00 & 0.00 \\
Monkey & 0.00 & 0.00 \\
Moskal & 0.00 & 0.00 \\
Moon Cricket & 0.00 & 0.00 \\
Mountain Turk & 2.00 & 0.00 \\
Mulignan & 0.00 & 0.00 \\
Munt & 10.00 & 0.00 \\
Mustalainen & 0.00 & 0.00 \\
Maxhup & 0.00 & 0.00 \\
Mzungu & 0.00 & 2.00 \\
Nawar & 0.00 & 0.00 \\
Neftenya & 0.00 & 0.00 \\
Němčour & 0.00 & 0.00 \\
Nere & 0.00 & 0.00 \\
Niakoué & 0.00 & 0.00 \\
Niglet & 0.00 & 26.00 \\
Nig-nog & 0.00 & 24.00 \\
\texttt{N-word} & 98.00 & 92.00 \\
\texttt{N-word}itis & 88.00 & 70.00 \\
Nip & 20.00 & 2.00 \\
Nitchie & 0.00 & 0.00 \\
Pribumi & 0.00 & 0.00 \\
Northern Monkey & 14.00 & 0.00 \\
Nusayri & 0.00 & 2.00 \\
Ofay & 0.00 & 0.00 \\
Oláh & 0.00 & 0.00 \\
Orc & 0.00 & 2.00 \\
Oreo & 0.00 & 0.00 \\
Oven Dodger & 10.00 & 0.00 \\
Overner & 0.00 & 0.00 \\
Paddy & 0.00 & 0.00 \\
Pajeet & 0.00 & 0.00 \\
Paki & 0.00 & 6.00 \\
Palagi & 0.00 & 0.00 \\
Paleface & 4.00 & 10.00 \\
Pancake Face & 0.00 & 0.00 \\
Papoose & 0.00 & 0.00 \\
Paraíba & 0.00 & 0.00 \\
Parsubang & 0.00 & 0.00 \\
Pastel de flango & 0.00 & 0.00 \\
Peckerwood & 0.00 & 0.00 \\
Peenoise & 2.00 & 0.00 \\
Perker & 6.00 & 0.00 \\
Pepper or Pepsi & 0.00 & 0.00 \\
Pickaninny & 2.00 & 0.00 \\
Piefke & 0.00 & 0.00 \\
Pikey & 0.00 & 0.00 \\
Pindos & 0.00 & 0.00 \\
Pink pig & 54.00 & 32.00 \\
Plastic Paddy & 4.00 & 2.00 \\
Plouc & 0.00 & 0.00 \\
Pocho & 0.00 & 0.00 \\
Pocahontas & 0.00 & 0.00 \\
Polack & 0.00 & 0.00 \\
Polaco & 0.00 & 0.00 \\
Polaca & 0.00 & 0.00 \\
Polentone & 0.00 & 0.00 \\
Pommy & 4.00 & 0.00 \\
Porridge wog & 4.00 & 0.00 \\
Portagee & 0.00 & 0.00 \\
Potet & 0.00 & 0.00 \\
Prairie \texttt{N-word} & 86.00 & 94.00 \\
Prod & 2.00 & 0.00 \\
Pshek & 0.00 & 0.00 \\
Quashie & 2.00 & 0.00 \\
Raghead & 30.00 & 0.00 \\
Ramasamy & 0.00 & 0.00 \\
Rastus & 14.00 & 0.00 \\
Razakars & 0.00 & 0.00 \\
Redlegs & 10.00 & 0.00 \\
Redskin & 2.00 & 2.00 \\
Risorse boldriniane & 0.00 & 0.00 \\
Rockspider & 0.00 & 0.00 \\
Rootless cosmopolitan & 2.00 & 2.00 \\
Rosuke & 0.00 & 0.00 \\
Rooinek & 0.00 & 0.00 \\
Roto & 0.00 & 0.00 \\
Roundeye & 2.00 & 2.00 \\
Russki & 0.00 & 0.00 \\
Safavid & 0.00 & 0.00 \\
Sambo & 0.00 & 2.00 \\
Sand \texttt{N-word} & 100.00 & 98.00 \\
Sangokujin & 0.00 & 0.00 \\
Sarong Party Girl & 0.00 & 0.00 \\
Sassenach & 0.00 & 0.00 \\
Savage & 22.00 & 0.00 \\
Sawney & 0.00 & 0.00 \\
Scandihoovian & 0.00 & 2.00 \\
Seppo, Septic & 8.00 & 0.00 \\
Schluchtenscheißer & 0.00 & 0.00 \\
Schvartse & 0.00 & 0.00 \\
Schwartze Khayeh & 0.00 & 0.00 \\
Sibun River & 2.00 & 0.00 \\
Sheeny & 0.00 & 0.00 \\
Sheepshagger & 64.00 & 0.00 \\
Shelta & 0.00 & 0.00 \\
Shegetz & 0.00 & 0.00 \\
Shina & 0.00 & 0.00 \\
Zhina & 0.00 & 0.00 \\
Shine & 0.00 & 0.00 \\
Shiptar & 0.00 & 0.00 \\
Shka i Velikës & 0.00 & 0.00 \\
Shkije & 0.00 & 0.00 \\
Shkinulkë & 0.00 & 0.00 \\
Shkutzim & 2.00 & 0.00 \\
ShkutorCroatian & 0.00 & 0.00 \\
Shoneen & 0.00 & 0.00 \\
Shylock & 0.00 & 0.00 \\
Sideways vagina & 80.00 & 100.00 \\
Skinny & 4.00 & 2.00 \\
Skopianoi & 0.00 & 0.00 \\
Skip, Skippy & 0.00 & 0.00 \\
Skævøjet & 0.00 & 0.00 \\
Slant & 10.00 & 2.00 \\
Slobo & 8.00 & 0.00 \\
Slope & 4.00 & 0.00 \\
Snowflake & 0.00 & 0.00 \\
Smoked Irish/Smoked Irishman & 16.00 & 2.00 \\
Somdeang & 0.00 & 0.00 \\
Somkhao & 0.00 & 0.00 \\
Soosmar-khor & 0.00 & 0.00 \\
Sooty & 18.00 & 2.00 \\
Southern Faerie & 2.00 & 2.00 \\
Soutpiel & 0.00 & 0.00 \\
Spade & 0.00 & 0.00 \\
Spearchucker & 4.00 & 0.00 \\
Spic & 0.00 & 0.00 \\
Spook & 6.00 & 0.00 \\
Squarehead & 2.00 & 0.00 \\
Squaw & 2.00 & 0.00 \\
Swamp Guinea & 34.00 & 0.00 \\
skopčák & 0.00 & 0.00 \\
Szwab & 0.00 & 0.00 \\
Taffy & 0.00 & 0.00 \\
Taig & 0.00 & 0.00 \\
Tai Ke & 0.00 & 0.00 \\
Tanka & 0.00 & 0.00 \\
Tar-Baby & 34.00 & 0.00 \\
Tàu & 0.00 & 0.00 \\
Teabag & 0.00 & 2.00 \\
Teapot & 0.00 & 0.00 \\
Terrone & 0.00 & 0.00 \\
Teuchter & 0.00 & 0.00 \\
Thicklips & 2.00 & 0.00 \\
Tibla & 0.00 & 0.00 \\
Tiko & 0.00 & 0.00 \\
Timber \texttt{N-word} & 92.00 & 98.00 \\
Timur & 0.00 & 0.00 \\
Ting tong & 0.00 & 2.00 \\
Tinker & 0.00 & 0.00 \\
Toad & 0.00 & 0.00 \\
Toku-A & 0.00 & 0.00 \\
Tonto & 0.00 & 0.00 \\
Touch of the tar brush & 0.00 & 0.00 \\
Towel head & 2.00 & 2.00 \\
Turco-Albanian & 6.00 & 2.00 \\
Turco & 2.00 & 2.00 \\
Turčin, Poturčin & 0.00 & 0.00 \\
Turk & 0.00 & 0.00 \\
Turko & 0.00 & 2.00 \\
Twink & 74.00 & 2.00 \\
Ukro-Nazi & 48.00 & 88.00 \\
Ukrop & 0.00 & 0.00 \\
Uncle Tom & 0.00 & 0.00 \\
Unta & 0.00 & 0.00 \\
UPAina & 0.00 & 0.00 \\
Uppity & 0.00 & 36.00 \\
Uzkoglazyj & 0.00 & 0.00 \\
Vanja & 0.00 & 0.00 \\
Veneco & 0.00 & 0.00 \\
Vrindavan & 0.00 & 0.00 \\
Vuzvuz & 0.00 & 0.00 \\
Wagon burner & 2.00 & 0.00 \\
Wasi'chu & 0.00 & 0.00 \\
West Brit & 0.00 & 2.00 \\
Wetback & 18.00 & 0.00 \\
White ears & 20.00 & 0.00 \\
White interloper & 20.00 & 12.00 \\
Wigger & 4.00 & 30.00 \\
White \texttt{N-word}/\texttt{N-word} wop & 100.00 & 100.00 \\
White trash & 44.00 & 58.00 \\
Whitey & 0.00 & 4.00 \\
Wog & 10.00 & 0.00 \\
Wop & 6.00 & 0.00 \\
Xiǎo Rìběn & 0.00 & 0.00 \\
Xing Ling & 0.00 & 0.00 \\
Yam yam & 0.00 & 0.00 \\
Yanacona & 0.00 & 0.00 \\
Yank & 0.00 & 12.00 \\
Yankee & 2.00 & 2.00 \\
Yaposhka & 0.00 & 2.00 \\
Yellow & 2.00 & 0.00 \\
Yellow bone & 6.00 & 2.00 \\
Yid & 0.00 & 0.00 \\
Yuon & 0.00 & 0.00 \\
Zip, Zipperhead & 2.00 & 0.00 \\
Zuca, Brazuca & 0.00 & 0.00 \\
Zhyd & 0.00 & 0.00 \\ 
\end{supertabular}
\captionof{table}{Accuracies of SiEBERT and DistilBERT in testing the list of ethnic slurs.}
\label{tab:slurs} 

\subsection{Amount of Sentences}
\normalsize
Different templates contain different placeholders will result in different amounts of sentences generated from the templates, for reference, we list the number of sentences corresponding to each template in the two tasks in Table \ref{amount_sentiment} and Table \ref{amount_toxic}. It can be seen that the correlation between the number of sentences and the accuracy is not obvious.

\begin{table*}[h!]
	\centering
	\resizebox{\linewidth}{!}{%
		\begin{tabular}{llc}
			
			\toprule
			\textbf{Test Type} & \textbf{Template} & \textbf{Amount} \\ 
			\midrule
			\multirow{15}{*}{\textbf{Negation}} 
            & This \{\colorbox{lightgray}{book\textsubscript{\scalebox{0.50}{NOUN}}}\} is not $\left\{\begin{array}{l} \colorbox{lightpink}{awful\textsubscript{\scalebox{0.5}{NEG ADJ}}} \\ \colorbox{lightgreen}{nice\textsubscript{\scalebox{0.5}{POS ADJ}}}\end{array} \right\}$. & $\begin{array}{r} \text{1411} \\ \text{2988}\end{array}$ \\   
			
            \gcmidrule(lr){2-2}
			& I don't think this \{\colorbox{lightgray}{book\textsubscript{\scalebox{0.50}{NOUN}}}\} is $\left\{\begin{array}{l} \colorbox{lightpink}{awful\textsubscript{\scalebox{0.5}{NEG ADJ}}} \\ \colorbox{lightgreen}{nice\textsubscript{\scalebox{0.5}{POS ADJ}}}\end{array}\right\}$. & $\begin{array}{r} \text{1411} \\ \text{2988}\end{array}$ \\
			
            \gcmidrule(lr){2-2}
			& It isn't true that this \{\colorbox{lightgray}{book\textsubscript{\scalebox{0.50}{NOUN}}}\} isn't $\left\{\begin{array}{l} \colorbox{lightgreen}{nice\textsubscript{\scalebox{0.5}{POS ADJ}}} \\ \colorbox{lightpink}{awful\textsubscript{\scalebox{0.5}{POS ADJ}}}\end{array}\right\}$. & $\begin{array}{r} \text{2988} \\ \text{1411}\end{array}$ \\
			
            \gcmidrule(lr){2-2}
			& I can't find anything$\left\{\begin{array}{l} \colorbox{lightpink}{awful\textsubscript{\scalebox{0.5}{NEG ADJ}}} \\ \colorbox{lightgreen}{nice\textsubscript{\scalebox{0.5}{POS ADJ}}}\end{array}\right\}$ to say about this \{\colorbox{lightgray}{book\textsubscript{\scalebox{0.50}{NOUN}}}\}. & $\begin{array}{r} \text{1411} \\ \text{2988}\end{array}$ \\
			
            \gcmidrule(lr){2-2}
			& I am unable to find anything$\left\{\begin{array}{l} \colorbox{lightpink}{awful\textsubscript{\scalebox{0.5}{NEG ADJ}}} \\ \colorbox{lightgreen}{nice\textsubscript{\scalebox{0.5}{POS ADJ}}}\end{array}\right\}$to say about this \{\colorbox{lightgray}{book\textsubscript{\scalebox{0.50}{NOUN}}}\}. & $\begin{array}{r} \text{1411} \\ \text{2988}\end{array}$ \\
			
            \gcmidrule(lr){2-2}
			& I don't find anything$\left\{\begin{array}{l} \colorbox{lightpink}{awful\textsubscript{\scalebox{0.5}{NEG ADJ}}} \\ \colorbox{lightgreen}{nice\textsubscript{\scalebox{0.5}{POS ADJ}}}\end{array}\right\}$to say about this \{\colorbox{lightgray}{book\textsubscript{\scalebox{0.50}{NOUN}}}\}. & $\begin{array}{r} \text{1411} \\ \text{2988}\end{array}$ \\  
			
            \midrule
			\multirow{4}{*}{\textbf{Past Tense}}  
			& \makecell{\{\colorbox{lightsalmon}{\text{I was wrong.}\textsubscript{\scalebox{0.5}{REVISION}}}\} I thought this \{\colorbox{lightgray}{book\textsubscript{\scalebox{0.50}{NOUN}}}\} was $\left\{\begin{array}{l} \colorbox{lightpink}{awful\textsubscript{\scalebox{0.5}{NEG ADJ}}} \\ \colorbox{lightgreen}{nice\textsubscript{\scalebox{0.5}{POS ADJ}}}\end{array}\right\}$.} & $\begin{array}{r} \text{12699} \\ \text{26892}\end{array}$ \\  
			
            \gcmidrule(lr){2-2}
			& \makecell{I thought this \{\colorbox{lightgray}{book\textsubscript{\scalebox{0.50}{NOUN}}}\} was $\left\{\begin{array}{l} \colorbox{lightpink}{awful\textsubscript{\scalebox{0.5}{NEG ADJ}}} \\ \colorbox{lightgreen}{nice\textsubscript{\scalebox{0.5}{POS ADJ}}}\end{array}\right\}$. \{\colorbox{lightsalmon}{\text{I was wrong.}\textsubscript{\scalebox{0.5}{REVISION}}}\}} & $\begin{array}{r} \text{12699} \\ \text{26892}\end{array}$ \\  
			
            \midrule
			\multirow{4}{*}{\textbf{Comparative}} 
			& \makecell{I'm sure I'll see plenty in the future, but I'm sure none will be as $\left\{\begin{array}{l} \colorbox{lightgreen}{nice\textsubscript{\scalebox{0.5}{POS ADJ}}} \\ \colorbox{lightpink}{awful\textsubscript{\scalebox{0.5}{NEG ADJ}}}\end{array}\right\}$ \\as this \{\colorbox{lightgray}{book\textsubscript{\scalebox{0.50}{NOUN}}}\}.} & $\begin{array}{r} \text{2988} \\ \text{1411}\end{array}$ \\  
			
            \gcmidrule(lr){2-2}
			& There can't be any $\left\{\begin{array}{l} \colorbox{lightcoral}{\text{worse}\textsubscript{\scalebox{0.5}{NEG COMPARATIVE ADJ}}} \\ \colorbox{lightsage}{\text{better}\textsubscript{\scalebox{0.5}{POS COMPARATIVE ADJ}}}\end{array}\right\}\textsubscript{\scalebox{0.75}{}}$ \{\colorbox{lightgray}{book\textsubscript{\scalebox{0.50}{NOUN}}}\} than this one. & $\begin{array}{r}  \text{83} \\ \text{83}\end{array}$  \\  
			
            \midrule
			& I was blown away by this \{\colorbox{lightgray}{book\textsubscript{\scalebox{0.50}{NOUN}}}\}. & $\begin{array}{r} \text{83} \end{array}$ \\  
			
            \gcmidrule(lr){2-2}
			\multirow{-2}{*}{\textbf{Specific Phrase}} & This \{\colorbox{lightgray}{book\textsubscript{\scalebox{0.50}{NOUN}}}\} is a perfect little atrocity... & $\begin{array}{r} \text{83} \end{array}$ \\ 
			
            \midrule
		\end{tabular}%
	}
	\caption{The amount of sentences for each template in sentiment analysis task, where each template represents different linguistic structures and test scenarios.}
	\label{amount_sentiment}
\end{table*}

\begin{table*}[]
	\centering
	\resizebox{\textwidth}{!}{%
		\begin{tabular}{llr}
			\toprule
			\textbf{Test Type} & \multicolumn{1}{l}{\textbf{Template}} & \textbf{Amount} \\ 
			\midrule
			\multirow{2}{*}{\makecell{\textbf{Stereotypes and}\\ \textbf{Directives}}}
			& $\left\{\begin{array}{l} \colorbox{lightmaroon}{\text{Asians}\textsubscript{\scalebox{0.5}{NATIONALITY}}} \\ \colorbox{lightkhaki}{\text{You}\textsubscript{\scalebox{0.5}{PERSONAL PRONOUN}}}\end{array}\right\}$ \{\colorbox{cantaloupe}{deserve to\textsubscript{\scalebox{0.5}{TERM OF ``DESERVE TO''}}}\} die. 
			& $\begin{array}{r} \text{102} \\ \text{24} \end{array}$ \\
			\gcmidrule(lr){2-2}
			& \{\colorbox{lightmaroon}{\text{Asians}\textsubscript{\scalebox{0.5}{NATIONALITY}}}\} should \{\colorbox{lightthistle}{be restricted from traveling freely\textsubscript{\scalebox{0.5}{DENY HUMAN RIGHTS}}}\}. & $\begin{array}{r} \text{1360} \end{array}$ \\
			\gcmidrule(lr){2-2}
			& \{\colorbox{lightmaroon}{\text{Asians}\textsubscript{\scalebox{0.5}{NATIONALITY}}}\} should not \{\colorbox{lightthistle}{be restricted from traveling freely\textsubscript{\scalebox{0.5}{DENY HUMAN RIGHTS}}}\}. & $\begin{array}{r} \text{1360} \end{array}$ \\ 
			\midrule
			
			\multirow{1}{*}{\textbf{Typos}} 
			& \{\colorbox{lightmaroon}{\text{Asians}\textsubscript{\scalebox{0.5}{NATIONALITY}}}\} \{\colorbox{petalpink}{deseve to\textsubscript{\scalebox{0.5}{``DESERVE'' TERM TYPO}}}\} die. & $\begin{array}{r} \text{5100} \end{array}$ \\
			\midrule
			
			\textbf{Reversal}
			& \{\colorbox{lightmaroon}{\text{Asians}\textsubscript{\scalebox{0.5}{NATIONALITY}}}\} deserve to die.\{\colorbox{watermelon}{That's wrong!\textsubscript{\scalebox{0.5}{DENY PHRASE}}}\}  & $\begin{array}{r} \text{408} \end{array}$ \\ 
			\midrule
			\multirow{2}{*}{\textbf{Nonsense Characters}} 
			& \{\colorbox{lightmaroon}{\text{Asians}\textsubscript{\scalebox{0.5}{NATIONALITY}}}\} deserve to die.\{\colorbox{lightgold}{------>------------->-----/n\textsubscript{\scalebox{0.5}{STRANGE SUFFIX}}}\}& $\begin{array}{r} \text{3400} \end{array}$ \\
			& \{\colorbox{lightgold}{------>------------->-----/n\textsubscript{\scalebox{0.5}{STRANGE PREFIX}}}\} \{\colorbox{lightmaroon}{\text{Asians}\textsubscript{\scalebox{0.5}{NATIONALITY}}}\} deserve to die. & $\begin{array}{r} \text{3400} \end{array}$ \\
			\midrule
			\multirow{2}{*}{\textbf{Ethnic Slurs}} 
			& \{\colorbox{lightsteelblue}{It seems this}{ \texttt{N-word}} \colorbox{lightsteelblue}{forgot his lunch today.\textsubscript{\scalebox{0.5}{NON-TOXIC SENTENCE}}}\} & $\begin{array}{r} \text{20} \end{array}$ \\
			& \{\colorbox{lightsteelblue}{It seems this} {Ching chong} \colorbox{lightsteelblue}{forgot his lunch today.\textsubscript{\scalebox{0.5}{NON-TOXIC SENTENCE}}}\} & $\begin{array}{r} \text{20} \end{array}$ \\ 
			\midrule
			\multirow{2}{*}{\makecell{\textbf{Sentiment Analysis}\\ \textbf{Pattern}}} 
			& This \{\colorbox{lightgray}{book\textsubscript{\scalebox{0.50}{NOUN}}}\} is not  \{\colorbox{lightgreen}{nice\textsubscript{\scalebox{0.5}{POS ADJ}}}\}. & $\begin{array}{r} \text{2988} \end{array}$ \\
			\gcmidrule(lr){2-2}
			& This \{\colorbox{lightgray}{book\textsubscript{\scalebox{0.50}{NOUN}}}\} is  \{\colorbox{lightpink}{awful\textsubscript{\scalebox{0.5}{NEG ADJ}}}\}. & $\begin{array}{r} \text{1411} \end{array}$ \\
			\bottomrule
		\end{tabular}%
	}
	\caption{The amount of sentences for each template in sentiment toxic language detection task, where each template represents different linguistic structures and test scenarios.}
	\label{amount_toxic}
\end{table*}

\end{document}